\documentclass[sigconf]{acmart} 
\usepackage[utf8]{inputenc}
\usepackage{here}
\usepackage{bmpsize}
\usepackage{enumerate}		
\usepackage{booktabs} 
\usepackage{amsmath} % AMS
\usepackage{amsthm}
\usepackage{algorithm}
\usepackage{algpseudocode}
\usepackage{cleveref}
\usepackage{xcolor}
\usepackage{latexsym}
\usepackage{listings}
\usepackage{mathtools}       % many extension of amsmath
\usepackage{stmaryrd}        % \llbracket, \rrbracket
\usepackage{bm}              % bold font
\usepackage{subcaption}

\usepackage{xcolor}
\usepackage{ifthen}
\usepackage{comment}

\usepackage[]{hyperref}

\providecommand{\argmax}{\operatornamewithlimits{argmax}} % argmax
\providecommand{\argmin}{\operatornamewithlimits{argmin}} % argmin
\DeclareMathOperator{\Tr}{Tr}     % trace
\DeclareMathOperator{\diag}{diag} % diagonal
\providecommand{\R}{\mathbb{R}} % real field
\providecommand{\E}{\mathbb{E}} % expectation
\providecommand{\T}{\mathrm{T}} % transpose
\renewcommand{\geq}{\geqslant} % geq
\renewcommand{\leq}{\leqslant} % leq
\DeclarePairedDelimiterX{\inner}[2]{\langle}{\rangle}{#1, #2}
\DeclarePairedDelimiter{\norm}{\lVert}{\rVert}
\DeclarePairedDelimiter{\abs}{\lvert}{\rvert}

\DeclareMathOperator{\sign}{sign}

\providecommand{\amacma}{\texttt{CMA-ES~with~WRA}}
\providecommand{\wra}{\texttt{WRA}}
\providecommand{\acma}{\texttt{Adversarial-CMA-ES}}
\providecommand{\zopgda}{\texttt{ZOPGDA}}

\copyrightyear{2022} 
\acmYear{2022} 
\setcopyright{acmlicensed}\acmConference[GECCO '22]{Genetic and Evolutionary Computation Conference}{July 9--13, 2022}{Boston, MA, USA}
\acmBooktitle{Genetic and Evolutionary Computation Conference (GECCO '22), July 9--13, 2022, Boston, MA, USA}
\acmPrice{15.00}
\acmDOI{10.1145/3512290.3528702}
\acmISBN{978-1-4503-9237-2/22/07}

\begin{document}
\title[CMA-ES with Worst-case Ranking Approximation]{Black-Box Min--Max Continuous Optimization Using CMA-ES with Worst-case Ranking Approximation}
%\title{Black-Box Min--Max Continuous Optimization Using Covariance Matrix Adaptation Evolution Strategy  \\ with Worst-case Ranking Approximation}

\author {Atsuhiro Miyagi}
\orcid{0000-0002-7729-8496}
\affiliation{%
 \institution{Taisei Corporation \& University of Tsukuba}
  % \streetaddress{1-1-1 Tennodai}
  % \city{Tsukuba} 
  % \state{Ibaraki} 
 % \postcode{305-8573}
}
\email{atsuhiro@bbo.cs.tsukuba.ac.jp}

\author{Kazuto Fukuchi}
\orcid{0000-0003-3895-219X}
\affiliation{%
 \institution{University of Tsukuba \& RIKEN AIP}
 % \streetaddress{1-1-1 Tennodai}
 % \city{Tsukuba} 
 % \state{Ibaraki} 
 % \postcode{305-8573}
}
\email{fukuchi@cs.tsukuba.ac.jp}

\author{Jun Sakuma}
\orcid{0000-0001-5015-3812}
\affiliation{%
 \institution{University of Tsukuba \& RIKEN AIP}  
 % \streetaddress{1-1-1 Tennodai}
 % \city{Tsukuba} 
 % \state{Ibaraki} 
 % \postcode{305-8573}
}
\email{jun@cs.tsukuba.ac.jp}

\author{Youhei Akimoto}
\orcid{0000-0003-2760-8123}
\affiliation{%
 \institution{University of Tsukuba \& RIKEN AIP}    
 % \streetaddress{1-1-1 Tennodai}
 % \city{Tsukuba} 
 % \state{Ibaraki} 
 % \postcode{305-8573}
}
\email{akimoto@cs.tsukuba.ac.jp}


\sloppy

\begin{abstract}
In this study, we investigate the problem of min--max continuous optimization in a black-box setting  $\min_{x} \max_{y}f(x,y)$.
A popular approach updates $x$ and $y$ simultaneously or alternatingly. 
However, two major limitations  have been reported in existing approaches.
(I) As the influence of the interaction term between $x$ and $y$ (e.g., $x^\T B y$) on the Lipschitz smooth and strongly convex--concave function $f$ increases, the approaches converge to an optimal solution at a slower rate.
(II) The approaches fail to converge if $f$ is not Lipschitz smooth and strongly convex--concave around the optimal solution.
To address these difficulties, we propose minimizing the worst-case objective function $F(x)=\max_{y}f(x,y)$ directly using the covariance matrix adaptation evolution strategy, in which the rankings of solution candidates are approximated by our proposed worst-case ranking approximation (\wra{}) mechanism.  
Compared with existing approaches, numerical experiments show two important findings regarding our proposed method.
(1) The proposed approach is efficient in terms of $f$-calls on a Lipschitz smooth and strongly convex--concave function with a large interaction term.
(2) The proposed approach can converge on functions that are not Lipschitz smooth and strongly convex--concave around the optimal solution, whereas existing approaches fail. 
\end{abstract}




\maketitle

\section{Introduction}\label{sec:intro}

\textbf{Background.}
Simulation-based optimization has been utilized in  various fields. In such optimizations, numerical simulations are used to evaluate the objective function on a solution candidate, with the conditions decided by preliminary investigation. 
For example, the coefficients in the governing equation or constitution rule should be decided beforehand when using the finite element method. However, these predetermined conditions contain  uncertainties in many cases. Hence, the numerical simulation results contain uncertainties \cite{Freitas2002, Wang20201}.
Therefore, finding a robust solution against these uncertainties is desirable for simulation-based optimization.

%To obtain a robust solution, one possible approach is to formulate a min--max optimization, given by 
To obtain a robust solution, previous studies for the automatic berthing control problem \cite{akimoto2021} and the electromagnetic scattering design problem \cite{Bertsimas2010} formulated a min--max optimization determined by following:
\begin{align}
	&\min_{x \in \mathbb{X}} \max_{y \in \mathbb{Y}} f(x, y) \enspace, \label{eq:minmax} 
\end{align}
where $f(x,y)$ is the objective function, $x \in \mathbb{X} \subseteq \R^m$ is a \emph{design variable}, and $y \in \mathbb{Y} \subseteq \R^n$ is an uncertain parameter that denotes the numerical simulation conditions, which is refered as the \emph{scenario variable} in this study. 
%We assume $y$ (we refer to this variable as the scenario variable in this paper) to be uncertain. 
A naive approach is to select the recommended scenario variable $y_\mathrm{est} \in \mathbb{Y}$ based on  expert judgment and obtain the following optimum solution:  $x_{y_\mathrm{est}}=\argmin_{x \in \mathbb{X}} f(x,y_\mathrm{est})$. 
However, because of the discrepancy between $y_\mathrm{est}$ and the scenario variable in a real environment $y_\mathrm{real}$, the performance of the solution $x_{y_\mathrm{est}}$ in the simulation does not guarantee satisfactory performance in the real-world environment. The main concern is that the solution may result in $f(x_{y_\mathrm{est}},y_\mathrm{real}) \gg f(x_{y_\mathrm{est}}, y_\mathrm{est})$.
Nevertheless, the optimal solution to \eqref{eq:minmax} guarantees the lower-bound of the performance in a real environment, i.e., $f(x,y_\mathrm{real}) \leq F(x) = \max_{y \in \mathbb{Y}}f(x,y)$ as long as $y_\mathrm{real} \in \mathbb{Y}$.
Therefore, the solution obtained by \eqref{eq:minmax} performs well in a real-world environment. When the appropriate $y_\mathrm{est}$ cannot be selected, it is important to consider the min--max problem shown in \eqref{eq:minmax}.

In this study, we consider min--max optimization \eqref{eq:minmax} with the following properties:
the gradient information of the objective function $f$ is unavailable (derivative-free optimization), and  
the objective function $f$ is not mathematically and explicitly expressed. Moreover, its characteristic constants, such as its Lipschitz constant, are unavailable (black-box optimization).
We refer to such a problem as the \emph{black-box min--max optimization} problem.

\textbf{Related works.}
%
%We first discuss two previous studies that were based on the derivative-free approach for the min--max optimization.
%
Liu et al.~\cite{Liu2020} proposed the zero-order projected gradient descent ascent (\zopgda{}), which searches the optimal solution using the approximated gradient of the objective function for $x$ and $y$. This approach updates $x$ and $y$ at iteration $t$ repeatedly as follows: 
\begin{align}
(x^{t+1}, y^{t+1}) = (x^t, y^t) + (-\eta_x \cdot B_x, \eta_y \cdot B_y) \enspace, \label{eq:xyupdate}
\end{align}
where $t$ is the number of iterations, $\eta_x$ and $\eta_y$ are the learning rates, and $B_x$ and $B_y$ are the approximated gradients $(\bar{\nabla}_{x} f, \bar{\nabla}_{y} f)$ of the objective function.
Numerical experiments showed that \zopgda{} is superior to STABLEOPT~\cite{Bogunovic2018}, in terms of the scalability of the optimization time against the problem dimension; STABLEOPT is a derivative-free approach based on Bayesian optimization. 
%However, the learning rate should be set according to the characteristics of the objective function, such as the maximum and the minimum eigenvalues of the Hessian of the objective function.

In another previous study \cite{akimoto2021}, the optimization approach \acma{} was proposed for the black-box min--max optimization. \acma{} updates $x$ and $y$ using \eqref{eq:xyupdate} with $B_x = \bar{x}^t - x^t$ and $B_y = \bar{y}^t - y^t$, where $\bar{x}^t$ and $\bar{y}^t$ are approximate solutions to $\argmin_{x \in \mathbb{X}}f(x, y^t)$ and $\argmax_{y \in \mathbb{Y}}f(x^t, y)$, respectively.
%In contrast to \zopgda{}, \acma{} is designed to adapt the learning rate $\eta=\eta_x=\eta_y$ in \eqref{eq:xyupdate} during the optimization.
Unlike \zopgda{}, which should set the learning rate according to the characteristics of the objective function, such as the maximum and minimum eigenvalues of the Hessian of the objective function, \acma{} is designed to adapt the learning rate $\eta=\eta_x=\eta_y$ in \eqref{eq:xyupdate} during the optimization.
Therefore, \acma{} should be more practical for black-box min--max optimization.
In numerical experiments, \acma{} was compared with co-evolutionary approaches \cite{Barbosa1999, Herrmann1999,Qiu2018}, which are also derivative-free approaches for black-box min--max optimization. \acma{} outperformed the existing co-evolutionary approaches in the worst-case scenario. It was observed for co-evolutionary approaches to fail to converge to the optimal solution, even on a strongly convex--concave and Lipschitz smooth (gradient is Lipschitz continuous) function.

Despite their promising results on some problems, \zopgda{} and \acma{} have limitations \cite{akimoto2021}.
\textbf{Difficulty (I)}: When the objective function is Lipschitz smooth and strongly convex--concave, the number of $f$-calls that \acma{} performs until it locates an $\epsilon$-optimal solution (a solution $x$ around the optimum solution $x^*$ satisfying $f(x) \leq f(x^*) + \epsilon$ for some $\epsilon > 0$) scales as $O\left( 1+\sigma_{\mathrm{max}}^2(H_{x,x}^{-1/2} H_{x, y} (-H_{y,y})^{-1/2}) \right)$, where $H_{x,x} = \nabla_x\nabla_x f(x^*, y^*)$, $H_{x, y} = \nabla_x\nabla_y f(x^*, y^*)$, and $H_{y,y} = \nabla_y\nabla_y f(x^*, y^*)$ are the blocks of the Hessian matrix of $f$ at the global min--max saddle point $(x^*, y^*)$; $\sigma_{\max}(\cdot)$ represents the maximum singular value. %, and $\epsilon$-optimal solution is a solution $x$ around the optimum solution $x^*$ and satisfies $f(x) \leq f(x^*) + \epsilon$ with positive value $\epsilon$.
In other words, the convergence slows down as the influence of the interaction term between $x$ and $y$, $H_{x, y}$ grows.
\textbf{Difficulty (II)}: \acma{} fails to converge to a min--max saddle point if the objective function is not a strongly convex--concave and Lipschitz  smooth function.
Although these issues have only been reported for \acma{}; similar limitations have been reported for the first-order approach \cite{Liang2019} on which \zopgda{} is based. In our experiments with \zopgda{}, these limitations were observed.
These situations occur frequently, and therefore are important issues that must be addressed.

In this study, we consider a black-box min--max optimization approach that can address the aforementioned  issues.

\textbf{Contribution.}
The study makes the following contributions: 

(1) A black-box min--max optimization approach, covariance matrix adaptation evolution strategy (CMA-ES) with the worst-case ranking approximation (\amacma{}), is proposed. 
\amacma{} aims to optimize the worst-case objective function $F$ using CMA-ES \cite{Hansen2001, Hansen2014, akimoto2019}, while the rankings of the worst-case objective function values of the solution candidates are approximated by the \wra{} mechanism. The \wra{} mechanism approximates the rankings of the solution candidates by solving the internal maximization problems approximately, $\max_{y} f(x, y)$, for each solution candidate using CMA-ES with a warm starting strategy and an early stopping strategy.

(2) To verify that \amacma{} can address Difficulty~(I),
we compared \amacma{} with the existing approaches, \zopgda{} and \acma, on a Lipschitz smooth and strongly convex--concave function. We empirically observed  that \amacma{} could locate an $\epsilon$-optimal solution within $O\left(\log(\sigma_{\mathrm{max}}(H_{x,x}^{-1/2} H_{x, y} (-H_{y,y})^{-1/2}))\right)$. We provide a theoretical but not rigorous reasoning for this scaling. 

(3) To verify that \amacma{} can address Difficulty~(II),
we conducted numerical experiments on functions that were not strongly convex--concave and Lipschitz smooth around the optimal solution. We compared \amacma{} with the existing approaches for these test problems. \amacma{} could locate an $\epsilon$-optimal solution, whereas the existing approaches failed.

(4) Additionally, we investigated how the number of $f$-calls performed  by \amacma{} changed if the coefficient of the interaction term, $H_{x, y}$, changed on functions that were not strongly convex--concave and Lipschitz smooth around the optimal solution. We observed a similar scaling of the number of $f$-calls to that of a Lipschitz smooth and strongly convex--concave function.

Our implementation of \amacma{} is publicly available.\footnote{\url{https://gist.github.com/a2hi6/ac511f101a494197b5fab56a407aa094}}

\section{Problem description}\label{sec:prob}

% In this section, we describe the assumptions of our target problem \eqref{eq:minmax} and the assumptions of the existing approaches.

Our objective is to find the optimal solution $x^*$ that minimizes the \emph{worst-case objective function} $F:\mathbb{X}\to\R$, defined as followings: 
\begin{align}
	F(x) = \max_{y \in \mathbb{Y}} f(x, y), \label{eq:Fx}
\end{align}
where $f:\mathbb{X}\times \mathbb{Y}\to \R$ is the objective function, and $\mathbb{X}\subseteq \R^m$ and $\mathbb{Y}\subseteq \R^n$ are the search domains for the design and scenario variables, respectively.
As mentioned in the introduction, we consider derivative-free and black-box situations. 
Therefore, the worst-case objective function $F$ is not explicitly available.

We introduce the definition of a \emph{min--max saddle point} of $f$ and the \emph{strong convexity--concavity}. A neighborhood of $z^* \in \mathbb{W} \subseteq \mathbb{R}^\ell$ is defined as a subset $\mathcal{E}_{z} \subseteq \mathbb{W}$, such that there exists an open ball $\mathbb{B}(z^*, r)=\{z \in \mathbb{W}:\norm{z-z^*} < r\}$ included in $\mathcal{E}_{z}$.
A critical point of $f$ is a point $(x, y)$, such that $\nabla f (x,y) = (\nabla f_x (x,y),\nabla f_y (x,y)) = 0$.

\begin{definition}[min--max saddle point \cite{akimoto2021}]
A point $(\tilde{x}, \tilde{y}) \in \mathbb{X} \times \mathbb{Y}$ is a local min--max saddle point of a function $f:\mathbb{X} \times \mathbb{Y} \rightarrow \mathbb{R}$, if there exists a neighborhood $\mathcal{E}_x \times \mathcal{E}_y \subseteq \mathbb{X} \times \mathbb{Y}$, including $(\tilde{x}, \tilde{y})$, such that for any $(x,y) \in \mathcal{E}_x \times \mathcal{E}_y$, the condition $ f(\tilde{x}, y) \leq f(\tilde{x}, \tilde{y}) \leq f(x,\tilde{y})$ holds. If $\mathcal{E}_x = \mathbb{X}$ and $\mathcal{E}_y = \mathbb{Y}$, the point $(\tilde{x}, \tilde{y})$ is called the global min--max saddle point. A strict min--max saddle point is one where the equality only holds if $(x,y) = (\tilde{x}, \tilde{y})$. A saddle point that is not a strict min--max saddle point is called a weak min--max saddle point.
\end{definition}
\begin{definition}[strongly convex concave function \cite{akimoto2021}]\label{def:strongcc}
A twice continuously differential function $f \in \mathcal{C}^2 ( \mathbb{R}^m \times \mathbb{R}^n, \mathbb{R})$ is locally $\mu$-strongly convex--concave around a critical point $(\check{x}, \check{y})$ for some $\mu > 0$ if there exist open sets $\mathcal{E}_x \subseteq \mathbb{R}^m$ including $\check{x}$ and $\mathcal{E}_y \subseteq \mathbb{R}^n$ including $\check{y}$, such that $H_{x,x}(x, y) - \mu\cdot  I$ and $-H_{y,y}(x, y) - \mu \cdot I$ are non-negative definite for all $(x,y) \in \mathcal{E}_x \times \mathcal{E}_y$. The function $f$ is a globally $\mu$-strongly convex--concave function if $\mathcal{E}_x = \mathbb{X}$ and $\mathcal{E}_y = \mathbb{Y}$. $f$ is locally or globally strongly convex--concave if it is locally or globally $\mu$-strongly convex--concave for some $\mu > 0$. 
\end{definition}

Importantly, if the objective function $f$ is twice-continuously differentiable and globally strongly convex--concave, there exists a unique critical point $(x^*, y^*) \in \R^{m} \times \R^{n}$, which is a unique min--max saddle point of $f$. Moreover, $x^*$ is a unique global minimum point of the worst-case objective function $F$. Let the \emph{worst-case scenario variables} $\hat{y}(x)$ for $x$ be defined as $\hat{y}(x) = \argmax_{y} f(x, y)$, i.e., $F(x) = f(x, \hat{y}(x))$. Then, it is known that $\hat{y}$ is uniquely determined, continuously differentiable, and $\hat{y}(x^*) = y^*$ \cite{akimoto2021}.

\section{Limitations of Existing Approaches} 

The existing approaches for the derivative-free min--max optimization problems, \zopgda{} \cite{Liu2020} and \acma{} \cite{akimoto2021}, are designed to converge to a local min--max saddle point of $f$ under the assumption that $f$ is at least locally strongly convex--concave. 
If the objective is locally strongly convex--concave, the simultaneous update of $x$ and $y$ of the form \eqref{eq:xyupdate} is intuitive. 
The reason for this is as follows. 
The worst-case scenario $\hat{y}(x^{t+1})$ is supposed to be close to $\hat{y}(x^{t})$ if $x^{t+1}$ and $x^{t}$ are close. 
If $y^{t}$ approximates $\hat{y}(x^{t})$ well, the next scenario $y^{t+1}$ that is close to $y^{t}$ is expected to approximate $\hat{y}(x^{t+1}) \approx \hat{y}(x^{t}) \approx y^{t}$. 
It has been demonstrated in \cite{akimoto2021} that this type of approach can converge linearly toward the global min--max saddle point if the learning rates $\eta_x$ and $\eta_y$ are sufficiently small.

However, as mentioned in the introduction, several limitations are highlighted in \cite{akimoto2021}. Among them, we focus on Difficulty (I) and (II), which have been introduced in \Cref{sec:intro}. Here, we elaborate on them with some examples.

{\bf Difficulty (I).} The learning rate must be sufficiently small for convergence, depending on the \emph{interaction term} between $x$ and $y$ of the objective function. For example, consider $f(x, y) = (a/2) x^2 + b x y - (c/2) y^2$. Then, \cite{akimoto2021} shows that the learning rate needs to be set proportional to $O(ac / (ac + b^2))$. As the coefficient of the interaction term, $b$, increases, compared with the coefficients of the non-interaction terms $a$ and $c$, the learning rate needs to be smaller. This results in a slower convergence, where the number of $f$-calls scales as $1 + b^2/(ac)$. A similar negative result was shown in \cite{Liang2019} for the first-order simultaneous update of $x$ and $y$. \zopgda{} is an approximation of the first-order counterpart; thus, the same limitation is expected and was observed in our experiments. 

{\bf Difficulty (II).} \acma{} reportedly fails to converge to a local min--max saddle point if $f$ is not strongly convex--concave and Lipschitz smooth (that is, the gradient is Lipschitz continuous) \cite{akimoto2021}. One such example is the  bi-linear function $f(x, y) = xy$ on a bounded domain $[-1, 1]\times[-1, 1]$. The failure of convergence of the first-order simultaneous update is also reported in \cite{Liang2019}. Therefore, \zopgda{} is also considered to fail to converge. 
Another example is $f(x, y) = (1/4) x^4 + b x y - (1/4) y^4$. Here, the situation is similar to that of $f(x, y) = (a/2) x^2 + b x y - (c/2) y^2$ but $a = x^2/2$ and $c = y^2/2$, i.e., $ac / (ac + b^2)$ becomes smaller as $x$ and $y$ approach $0$. Therefore, we expect the learning rate to be smaller as the algorithm approaches the global min--max saddle point, jeopardizing the linear convergence. For both examples, we observe that \acma{} and \zopgda{} fail to converge in our experiments.

%Another example is $f(x, y) = (1/4) x^2 + xy - (1/4) y^2$. For both examples, we observe that \acma{} and \zopgda{} fail to converge in our experiments. Another example is $f(x, y) = (1/4) x^2 + xy - (1/4) y^2$. In this case, the situation is similar to that of $f(x, y) = (a/2) x^2 + b x y - (c/2) y^2$ but $a = x^2/2$ and $c = y^2/2$. That is, $ac / (ac + b^2)$ becomes smaller as $x$ and $y$ approach $0$. Therefore, we expect that the learning rate needs to be smaller as the algorithm approaches the global min--max saddle point, which will jeopardize the linear convergence.

A possible cause of these limitations is the sensitivity of the worst-case scenarios $\hat{y}(x)$ against small changes in $x$. In the case of $f(x, y) = (a/2) x^2 + b x y - (c/2) y^2$, we have $\hat{y}(x) = (b / c) x$, i.e., the change in the worst-case scenario is proportional to $b/c$. If $b \gg c$, a small change in $x$ may lead to a great change in $\hat{y}(x)$. Then, the simultaneous update \eqref{eq:xyupdate} may fail to keep track of the worst-case scenario. To prevent this, the learning rate $\eta$ must be set sufficiently small, resulting in a slow convergence. In the case of $f(x, y) = x y$ on $[-1, 1]\times[-1, 1]$, the worst-case scenario is $\hat{y}(x) = \sign(x)$, which is not a continuous function of $x$ around $x^* = \argmin_{x} F(x) = 0$. A small change in $x$ near $x^* = 0$ could result in a sign flip for $x$, changing the worst-case scenario between $-1$ and $1$. Consequently, the simultaneous update \eqref{eq:xyupdate} may fail to keep track of the worst-case scenario. Thus, in this case, a small learning rate is ineffective.

\section{Proposed Approach}

We propose a novel approach to black-box min--max optimization problems \eqref{eq:minmax}, named the CMA-ES with the worst-case ranking approximation  (\amacma{}).
% \footnote{
% Algorithm is presented at https://gist.github.com/a2hi6/ac511f101a494197b5fab56a407aa094}
% ). 
This approach attempts to minimize the worst-case objective function $F$ directly using CMA-ES \cite{Hansen2001, Hansen2014, akimoto2019} to mitigate the aforementioned  limitations of the existing approaches. The worst-case objective function value $F(x)$ for each solution candidate $x$ is then approximated by solving the maximization problem $\max_{y} f(x, y)$. 
The proposed worst-case ranking approximation (\wra{}) mechanism uses a warm starting strategy and an early stopping strategy to reduce the number of $f$-calls for the internal maximization problems.

%The proposed worst-case ranking approximation (\wra{}) mechanism aims to reduce the number of $f$-calls for the internal maximization problems via a warm starting strategy and an early stopping technique.

\subsection{Addressing Difficulty (I) and (II)}
%\subsection{Addressing difficulties (I) and (II)}

Our main strategy to address Difficulty (I) and (II) described in the previous section is to minimize the worst-case objective function $F$ directly. Here, we explain why minimizing $F$ is expected to mitigate these difficulties. 

%Our main idea to address Difficulties (I) and (II) described in the previous section, which allows us to minimize the worst-case objective function $F$ directly. Here, we describe why minimizing $F$ is expected to mitigate these difficulties. 

First, we explain why we expect that minimizing $F$ will not suffer from a large interaction term (\textbf{Difficulty (I)}).
An arbitrary strongly convex--concave and Lipschitz smooth function can be approximated by a convex--concave quadratic function around the global min--max saddle point.
Therefore, for simplicity, consider a convex--concave quadratic function $f(x, y) = \frac12 x^\T A x + x^\T B y - \frac12 y^\T C y$, where $A \in \R^{m\times m}$ and $C \in \R^{n \times n}$ are symmetric positive definite, and $B \in \R^{m \times n}$ is an arbitrary matrix.
The worst-case scenario is $\hat{y}(x) = C^{-1} B^\T x$, and the worst-case objective function is $F(x) = f(x, \hat{y}(x)) = \frac12 x^\T \big(A + B C^{-1} B^\T \big) x$. 
Irrespective of the coefficients, this is a convex quadratic function. 
An approach exploiting the second-order information, such as CMA-ES, empirically shows linear convergence on an arbitrary convex quadratic function with a convergence rate independent of its Hessian matrix \cite{invariance}.
% can achieve the same convergence rate on an arbitrary convex quadratic function as that on the spherical function $\frac12 \norm{x}^2$ \textcolor{red}{because of its invariance properties \cite{invariance}}. 
%Obviously, the convergence rate on $\frac12 \norm{x}^2$ is irrelevant to the interaction term, $B$. 
Therefore, minimizing $F$ by CMA-ES is expected to show a linear convergence with a convergence rate independent of the interaction term.

Second, we show how the proposed approach can mitigate the issue of convergence on a function $f$ that is not strongly convex--concave and Lipschitz smooth (\textbf{Difficulty (II)}). This is, to some extent, intuitive. The proposed approach directly minimizes the worst-case objective $F$ ; thus, it will converge toward a local optimal solution of $F$ if $F$ is solvable by the search algorithm. For example, in the case of a bi-linear objective function $f(x, y) = x y$ on $[-1, 1]\times[-1, 1]$, the worst-case objective function is $F(x) = \abs{x}$. This is a monotonic transformation of a quadratic function $x^2$. A comparison-based search algorithm, such as CMA-ES, is known to be invariant to the monotonic transformation of the objective function. Therefore, if a comparison-based search algorithm that can solve a quadratic function efficiently is used, the worst-case objective function $F$ can also be efficiently solved.

\subsection{Worst-case Ranking Approximation}

The difficulty in directly solving the worst-case objective function $F$ is that we must evaluate $F(x)$ for each solution candidate $x$ by solving the maximization problem $\max_{y} f(x, y)$. The maximization problem cannot be solved precisely because $f$ is a black-box function. Hence, one must rely on a numerical solver. However, this is time--consuming because each $F(x)$ evaluation requires a single optimization process, which necessitates several $f$-calls. 

For each maximization problem, we use (a) warm starting and (b) early stopping of the numerical solver to reduce the number of $f$-calls.
The proposed approach uses CMA-ES as the numerical solver for the worst-case objective function $F$. At each iteration $t$, it samples $\lambda$ solution candidates, $x_1, \dots, x_{\lambda}$, from the Gaussian distribution $\mathcal{N}(m^t, \Sigma^t)$. Their rankings, denoted as $\mathrm{Rank}_F (\{x_i\}_{i=1}^{\lambda})$, are then computed based on $F$, which is now approximated by solving $\max_{y} f(x, y)$. The distribution parameters, mean vector $m$, covariance matrix $\Sigma$, and any other dynamic parameters $\theta$, are updated based on the solution candidates and their rankings. 
We have two important remarks. (1) CMA-ES \cite{Hansen2001, Hansen2014, akimoto2019} is comparison-based; thus, it behaves identically on $F$ and its approximation $\hat{F}$ if $\mathrm{Rank}_F (\{x_i\}_{i=1}^{\lambda})$ and $\mathrm{Rank}_{\hat{F}} (\{x_i\}_{i=1}^{\lambda})$ are equivalent. That is, $\hat{F}$ does not need to approximate $F$ more accurately than that required to approximate the rankings. 
This point is important in designing a stopping condition for the maximization problem.
(2) The search distribution $\mathcal{N}(m^t, \Sigma^t)$ does not significantly change in one iteration; therefore, the solution candidates generated in the current and last iteration are similarly distributed. Therefore, the worst-case scenario for the solution candidates generated in this iteration are expected to be distributed similarly to the solution candidates. This suggests the necessity of looking for the worst-case scenarios based on the result of previous iterations.

Therefore, we design the worst-case ranking approximation (\wra{}) mechanism. It takes $\lambda$ solution candidates as the input and returns their approximate rankings. To reduce $f$-calls inside \wra{}, the warm starting and early stopping strategies are implemented. The algorithm of \wra{} is summarized in \Cref{alg:wra}. 
It internally maintains $\lambda$ CMA-ES instances for the worst-case scenario search. The following sections cover the warm starting and early stopping strategies for these internal CMA-ES instances. 

\subsubsection{Warm Starting Strategy}\label{sec:initcmay}

For each solution candidate $x_i$, we select and run one of the internal CMA-ES to approximate $F(x_i)$. The purpose of the warm starting strategy is to help us choose an internal CMA-ES that will  reduce  the number of $f$-calls. 

The warm starting part is described in \Cref{line:initstart}--\Cref{line:initend} of \Cref{alg:wra}.
Let $y_k$ be the worst scenario parameter obtained by the $k$th CMA-ES instance in the last iteration. 
%Let $y_k$ be the scenario parameter that is the worst scenario parameter found by the $k$th CMA-ES instance in the last iteration. 
Then, for each solution candidate $x_i$ ($i = 1, \dots, \lambda$), we compute the objective function values $f(x_i, y_k)$ for $k =1, \dots, \lambda$. 
The worst-case scenario is then selected. Let $k_i^\mathrm{worst} = \argmax_{k} f(x_i, y_k)$ be the index of the worst-case scenario. 
Then, we select the $k_i^\mathrm{worst}$th CMA-ES instance for the worst-case scenario search for $x_i$. 
Starting with the CMA-ES instance that generates the worst-case scenario for $x_i$, we expect that the number of $f$-calls for the adaptation of the distribution parameters to be significantly lower, as compared to using a new CMA-ES with initial distribution parameters.\footnote{
CMA-ES has dynamic parameters, $\theta$, other than the distribution parameters, such as the so-called evolution paths. After each $\wra$ call, we only keep the solution $y_k$ and the distribution parameters $(m_k, \Sigma_k)$ and initialize all the other parameters, $\theta$, of each internal CMA-ES instance. That is, we avoid sharing the dynamic parameters $\theta$ for worst-case search for different solution candidates. This is to avoid a systematic bias caused by the change in the objective function because of the change in the solution candidate. The phenomenon is explained in \cite{akimoto2021}.
}
If the same CMA-ES instance is selected for a  different solution candidate, a clone is created.

\subsubsection{Early Stopping Strategy}\label{sec:stopcmay}

In a double loop strategy, determining the best time to stop the internal maximization process is difficult. However, as mentioned earlier, we can stop the worst-case scenario search without any compromise if the rankings, $\mathrm{Rank}_F (\{x_i\}_{i=1}^{\lambda})$, of the worst-case objective function values of solution candidates are correctly estimated.  Accordingly,  this condition is eased. We stop the worst-case scenario search if Kendall's rank correlation coefficient $\tau$ \cite{kendall} between the worst-case objective function values $\{F(x_i)\}_{i=1}^{\lambda}$ and its approximated values $\{\hat{F}(x_i)\}_{i=1}^{\lambda}$ is sufficiently high, for example, $\tau \geq \tau_\mathrm{threshold}$. 
Two CMA-ES with a high $\tau$ value should behave similar in each other \cite{Akimoto2022surrogate}; therefore, $\tau$ is frequently used to measure the quality of a surrogate function \cite{lq-cma,akimoto2019multi,miyagi2021}.
%A high $\tau$ value is expected to result in similar behavior \cite{Akimoto2022surrogate}; $\tau$ is often used to measure the quality of a surrogate function \cite{lq-cma,akimoto2019multi,miyagi2021}. 
However, because we cannot obtain $F(x_i)$, $\tau$ cannot be computed. Therefore, $\tau$ is approximated using the worst-case objective function values estimated in the current iteration and those estimated in previous iterations.

The worst-case ranking approximation with an early stopping strategy is described in \Cref{line:worststart}--\Cref{line:worstend} of \Cref{alg:wra}.
Let the first estimate of the worst-case objective function value for each solution candidate $x_i$ be denoted by $F_i^0$; then, all the CMA-ES instances are run for a certain number of iterations, which will be described later. We call it a \emph{round}, and the round is counted by $j \geq 1$.
The worst-case objective function value for each solution candidate $x_i$ estimated after the round $j$ is denoted by $F_i^j$. Then, $\tau$ between $\{F_i^j\}_{i=1}^{\lambda}$ and $\{F_i^{j-1}\}_{i=1}^{\lambda}$ is computed as $\tau$ between the ground truth worst-case objective function values $\{F(x_i)\}_{i=1}^{\lambda}$ and their estimates $\{F_i^{j-1}\}_{i=1}^{\lambda}$. 
A round is repeated until $\tau > \tau_\mathrm{threshold}$. 

In each round, each CMA-ES instance run is terminated if the worst-case scenario improves $c_\mathrm{max}$ times. Here, we assume that the worst-case scenario has been significantly improved over the last round. Additionally, the run is terminated if all the coordinate-wise standard deviation, $\sqrt{\Sigma_{\ell,\ell}}$, become smaller than the threshold $V_\mathrm{min}$. 
In this case, we assume that the search distribution has already converged, and the worst-case scenario will not be updated significantly anymore. However, if this condition is satisfied in the last call of \wra{} (that is, in the iteration of the CMA-ES solving $F$), this condition is immediately satisfied in the first round of the current call of \wra. To prevent this, we force the internal CMA-ES instance to run at least $T_{\mathrm{min}}$ iterations.

\subsubsection{Post processing}

After computing the worst-case ranking approximation, we perform post-processing (\Cref{line:poststart}--\Cref{line:postend} in \Cref{alg:wra}) for the next \wra{} call. 
First, we prevent the coordinate-wise standard deviation $\sqrt{[\Sigma_k]_{\ell,\ell}}$ from becoming smaller than $V_{\min}$. Otherwise, the termination condition in each round of the worst-case scenario search will be satisfied immediately after $T_{\min}$ iterations.
Second, we prevent the Gaussian distributions of $\lambda$ CMA-ES instances from converging to the same point. 
It is important to distance the CMA-ES instances from each other because the worst-case scenarios can be distinct, even for close solution candidates, for example, on a bi-linear function.
If the output worst-case scenarios of two CMA-ES instances are close to each other (the distance is smaller than $V_{\min} \cdot \sqrt{n}$), we reset the distribution parameters of one of these instances.
% $(m_k, \Sigma_k)=(m_k^0, \Sigma_k^0)$  and sample a new $y$ from the reset distribution, which $(m_k^0, \Sigma_k^0)$ are the given parameters at initial iteration.

\newcommand{\flgworstsearch}{\bm{1}_{\theta_{i}-\texttt{update}}}

\begin{algorithm}[t]
	\caption{Worst-case ranking approximation}\label{alg:wra}
    \begin{algorithmic}[1]\small
    \Require solution candidates to be ranked: $x_1, \dots, x_{\lambda}$ 
    \Require stopping conditions: $\tau_{\mathrm{threshold}} \in (0, 1]$, $c_{\max} \geq 1$, $V_{\min} \geq 0$
    \Require $(m_1, \Sigma_1), \dots, (m_{\lambda}, \Sigma_{\lambda})$ and $y_i \sim N(m_i, \Sigma_i)$ \text{ for  } $i=1,...,\lambda$
    \State\label{line:initstart} \texttt{// initialization part}
    \For{$i = 1, \dots, \lambda$}
    \State Evaluate $f(x_{i}, y_k)$ for all $k = 1,\dots,\lambda$
    \State Select the worst index $k^\mathrm{worst}_{i} = \argmax_{k \in \{ 1, \dots,\lambda\}} f(x_i, y_k)$
    and let $\tilde{y}_i = y_{k^\mathrm{worst}_{i}}$, $F^{0}_{i} = f(x_i, \tilde{y}_{i})$, $\tilde{m}_{i} = m_{k^\mathrm{worst}_{i}}$, $\tilde{\Sigma}_{i} = \Sigma_{k^\mathrm{worst}_{i}}$
    \State Initialize internal parameter $\theta_i$ of a CMA-ES instance
    \EndFor \label{line:initend}
    \State\label{line:worststart} \texttt{// worst-case ranking approximation part}
    \State $\tau = -1$, $j = 0$, $t_1, \dots, t_{\lambda} = 0$, $h_1, \dots, h_{\lambda} = 0$
    \While{ $\tau \leq \tau_{\mathrm{threshold}}$}
    \State $j = j + 1$
    \For {$i = 1, \dots, \lambda$}
    \State $c = 0$
    \While{$c < c_{\max}$ or $h_i = 0$}
    \State Perform one iteration of CMA-ES with $(\tilde{m}_i, \tilde{\Sigma}_i, \theta_{i})$ and obtain the worst candidate $y_i'$ and the updated parameters $(\tilde{m}_i, \tilde{\Sigma}_i, \theta_{i})$
    \If {$F^{j}_i < f(x^t_i,\tilde{y}_i')$}
    \State $F^j_i = f(x^t_i,\tilde{y}_i')$, $\tilde{y}_i = \tilde{y}_i'$, and $c = c + 1$
    \EndIf
    \State $h_i = 1$ \textbf{if} $\max_{\ell}\left\{\sqrt{[\tilde{\Sigma}_i]_{\ell,\ell}}\right\} < V_{\min}$ and $t_i \geq T_{\min}$
    \EndWhile
    \EndFor
   \State $\tau = \text{Kendall}(\{F^{j-1}_{i}\}_{i=1}^{\lambda}, \{F^{j}_{i}\}_{i=1}^{\lambda})$
    \EndWhile\label{line:worstend}
    \State \texttt{// post-process part}\label{line:poststart}
    \For{$i = 1, \dots, \lambda$}
    \State $(y_i, m_i, \Sigma_i) = (\tilde{y}_{i}, \tilde{m}_i, \tilde{\Sigma}_i)$    
    \State $D_i = \diag\left(\max\left(1, \frac{V_{\min}}{\sqrt{[\Sigma_i]_{1,1}}}\right), \dots, \max\left(1, \frac{V_{\min}}{\sqrt{[\Sigma_i]_{n,n}}}\right)\right)$
    \State $\Sigma_i = D_i \Sigma_i D_i$
    \For {$k= i+1, \dots, \lambda$}
    \If{$\text{distance}({y}_i, {y}_k) < V_{\min} \cdot \sqrt{n}$}
    \State Reset $(m_k, \Sigma_k)$ and sample $y_k \sim N(m_k, \Sigma_k)$
    \EndIf
    \EndFor
    \EndFor\label{line:postend}
\State \Return the rankings of $F_1^j, \dots, F_{\lambda}^j$
  \end{algorithmic}
\end{algorithm}

\section{Numerical experiments }

To test the following hypotheses, we compare \amacma{} with the existing approaches: \acma{}\footnote{https://gist.github.com/youheiakimoto/ab51e88c73baf68effd95b750100aad0} and
\zopgda{}\footnote{https://github.com/KaidiXu/ZO-minmax} in numerical experiments.
(a) \amacma{} is more efficient in terms of the number of $f$-calls if the objective function is Lipschitz smooth and strongly convex--concave, but the influence of the interaction term between $x$ and $y$ is large (\Cref{sec:ex1}).
(b) \amacma{} can converge to the optimal solution $x^*$ even if the objective function is not locally Lipschitz smooth and strongly convex--concave around $x^*$, whereas the existing approaches fail to converge (\Cref{sec:ex2}).
Additionally, we investigate how much the number of $f$-calls spent by \amacma{} scales when a coefficient $b$ of the interaction term $b x^T y$ is changed on objective functions that are not necessarily Lipschitz smooth and strongly convex--concave around $x^*$ (\Cref{sec:ex3}).

\begin{table*}[]
    \centering\small
    \caption{Test problem definitions and their characteristics. 
    The domains are $\mathbb{X} = [-3, 3]^m$ and $\mathbb{Y} = [-3, 3]^n$. The worst-case scenario for each solution $x \in \mathbb{X}$ is denoted by $\hat{y}(x) = (\hat{y}_1(x), \dots, \hat{y}_n(x))$. The characteristics are denoted as follows: strongly convex (st-cv), strongly concave (st-cc), convex (cv), concave (cc), smooth (sm), and non-smooth (non-sm).
    The scalar $b > 0$ is introduced to control the influence of the interaction term between $x$ and $y$. }
    \label{tab:testp}
    \begin{tabular}{l|c|c|l|l}
    \toprule
    $f$ & 
    $x$ &
    $y$ & 
    $\hat{y}_i(x)$ & 
    Optimum \\
    \midrule
    
    $f_1 = x^{T} y$ &
    linear & 
    linear &
    $3 \sign(x_i)$
    & 
    $x^* = 0$ \\

    $f_2 = \frac12 \norm{x}_2^2 + x^T y$ & 
    sm st-cv & 
    linear  & 
    $3 \sign(x_i)$ & 
    $x^* = 0$ \\

    $f_3 = \frac12 \norm{x+1}_2^2 + \frac{1}{10} x^T y$ & 
    sm st-cv & linear & 
    $3 \sign(x_i)$ &
    $x^*=-0.7$  \\

    $f_4 = \frac{1}{2} \norm{x}_2^2 + x^T y + \frac{1}{2} \norm{y}_2^2$ &
    sm st-cv & sm st-cv  &    
    $\begin{cases}
    3 \text{ or } -3 & x_i = 0 \\
    3 \sign(x_i) & 0 < \abs{x_i} \leq 3
    \end{cases}$ &
    $x^*=0$  \\
        
    $f_5 = \frac{1}{2} \norm{x}_2^2 + b x^T y - \frac{1}{2} \norm{y}_2^2$ &
    sm st-cv & sm st-cc  & 
    $\begin{cases}
    b x_i & \abs{x_i} \leq 3/b \\
    3 \sign(x_i) & \abs{x_i} > 3/b
    \end{cases}$ &
    $x^* = 0$ \\

    $f_6 = \frac12 \norm{x}_2^2 + \norm{x}_1 + b x^T y - \norm{y}_1 - \frac12 \norm{y}_2^2$ & 
    non-sm st-cv & non-sm st-cc & 
    $\begin{cases}
    0 & \abs{x_i} \leq 1/b \\
    bx_i - \sign(x_i)  & 1/b < \abs{x_i} \leq 4/b \\
    3 \sign(x_i) & 4/b < \abs{x_i}
    \end{cases}$ &
    $x^* = 0$  \\

    $f_7 = \frac14 \norm{x}_2^4 + b x^T y - \frac14\norm{y}_2^4$ &
    cv & cc & 
    $\begin{cases}
    (b/\norm{x}_2^2)^{1/3} x_i & (b/\norm{x}_2^2)^{1/3} \abs{x_i} \leq 3 \\
    3 \sign(x_i) & (b/\norm{x}_2^2)^{1/3} \abs{x_i} > 3
    \end{cases}$ &
    $x^* = 0$  \\

    $f_8 = \norm{x}_1 + b x^T y - \norm{y}_1$ &
    non-sm cv & non-sm cc &   
    $
    \begin{cases}
    0 & \abs{x_i} \leq 1/b \\
    3\sign(x_i) & \abs{x_i} > 1/b 
    \end{cases}
    $ &
    $x^* = 0$ \\
    \bottomrule
    \end{tabular}
\end{table*}

\subsection{Common Settings}

We designed eight test problems, summarized in \Cref{tab:testp}. 
They are designed to have different characteristics (smoothness, convexity,  and concavity) around the optimal solution of the objective function. 
The search domains of the design variables and scenario variables are $\mathbb{X}=[-3, 3]^{m}$ and $\mathbb{Y}=[-3, 3]^{n}$, respectively. 
The dimension of the design variables is $m=20$, and the dimension of the scenario variables is $n=20$. 

The configuration of \amacma{} is as follows:
The hyperparameters for \wra{} are set as follow: $\tau_{\rm threshold} = 0.7$, $c_{\max} = 2$, $V_{\min} = 10^{-4}$, and $T_{\min} = 10$.
The initial mean vectors and the covariance matrices of the internal CMA-ES instances are $m_{i} \sim \mathcal{U}(-3, 3)^n$ and $\Sigma_{i} = \diag(1.5, \dots, 1.5)^2$. 
When the distribution parameters of the internal CMA-ES instances are reset during the post processing phase of \wra, we use the same initialization. 
In the CMA-ES solving $F$, the initial mean vector is drawn from $m^0 \sim \mathcal{U}(-3, 3)^m$, and the initial covariance matrix is set to $\Sigma^0 = \diag(1.5, \dots, 1.5)^2$. 
The hyperparameters and the initial values of the dynamic parameters $\theta$ for the CMA-ES instances in \wra{} and the CMA-ES solving $F$ are set to their default values, as proposed in~\cite{akimoto2019}.

The hyperparameters for \zopgda{} and \acma{} are set based on the original studies \cite{Liu2020} and \cite{akimoto2021}, respectively.
For \zopgda{}, the learning rate parameters were set to  $\eta_x=0.02$ and $\eta_y=0.05$.
For \acma{}, we set $G_\mathrm{tol} = 10^{-7}$. The distribution parameters are initialized in the same way as \amacma.

To deal with the box constraint, \zopgda{} by default uses the projected gradient. 
\acma{} and \amacma{} use the mirroring technique along with upper-bounding of the coordinate-wise standard deviation \cite{yamaguchi2018}.

We evaluate the performance of each optimization algorithm by running $20$ independent runs. 
The termination criteria are as follows.
The number of $f$-calls in each run is limited to 
%$50,000,000$
$20,000,000$. 
Before the number of $f$-calls reaches $20,000,000$, the run is considered a success if $\abs{F(m^t) - F(x^*)}\leq 10^{-6}$ is satisfied. For \zopgda, $m^t$ is considered the estimate $x^t$ of the solution at iteration $t$.

\subsection{Experiment 1}\label{sec:ex1}

To verify hypothesis (a), we applied three approaches to $f_5$ at $b\in\{1,3,10,30,100\}$.   

The results of the experiment are shown in \Cref{fig:ex1}. As shown in \Cref{fig:ex1}, \amacma{} improved the scalability regarding the number of $f$-calls until convergence at coefficient $b$ compared with \zopgda{} and \acma{}. The increment for the number of $f$-calls with $b$ was approximately proportional to $\log(b)$. This result will be discussed in \Cref{sec:discussion}. In this experiment, when $10 \geq b$, the number of $f$-calls performed by \amacma{} was less than that by the others. 

We consider the results of \acma{} and \zopgda{}. First, the number of $f$-calls increased proportional to $b^2$ when $b=\{1,3\}$.
At small $b$, the existing approaches converged with less $f$-calls than \amacma{}.
For $b=10$, we expected from the fitted curve in \Cref{fig:ex1} that \acma{} converges successfully within the $f$-calls budget. However, it failed.
This was probably because of the box constraint. 
The theoretical analysis in \cite{akimoto2021} assumes unbounded domains.
Under the box constraint in this experiment, the character of $f_5$ at $b=10$ resembled that of $f_1$. Concretely, when a design variable is in $\abs{x_i} > 0.3 = 3 / b$ for each $i$, the $i$th coordinate of the worst scenario is $\hat{y}_i(x) = 3 \sign(x_i)$, which is the same as the worst scenario on $f_1$. As we will see in the next experiment, \acma{} fails to converge to the optimal solution. 
Therefore, we believe that \acma{} had difficulty converging toward the area with $\abs{x_i} \leq 0.3$ for all $i = 1,\dots,m$.
Second, \zopgda{} could not converge to the optimal solution in any trials where $10 \geq b$ because the learning rate was not tuned. 

\providecommand{\exss}{0.333}
\begin{figure}[t]
\centering
    \includegraphics[width=\exss\textwidth]{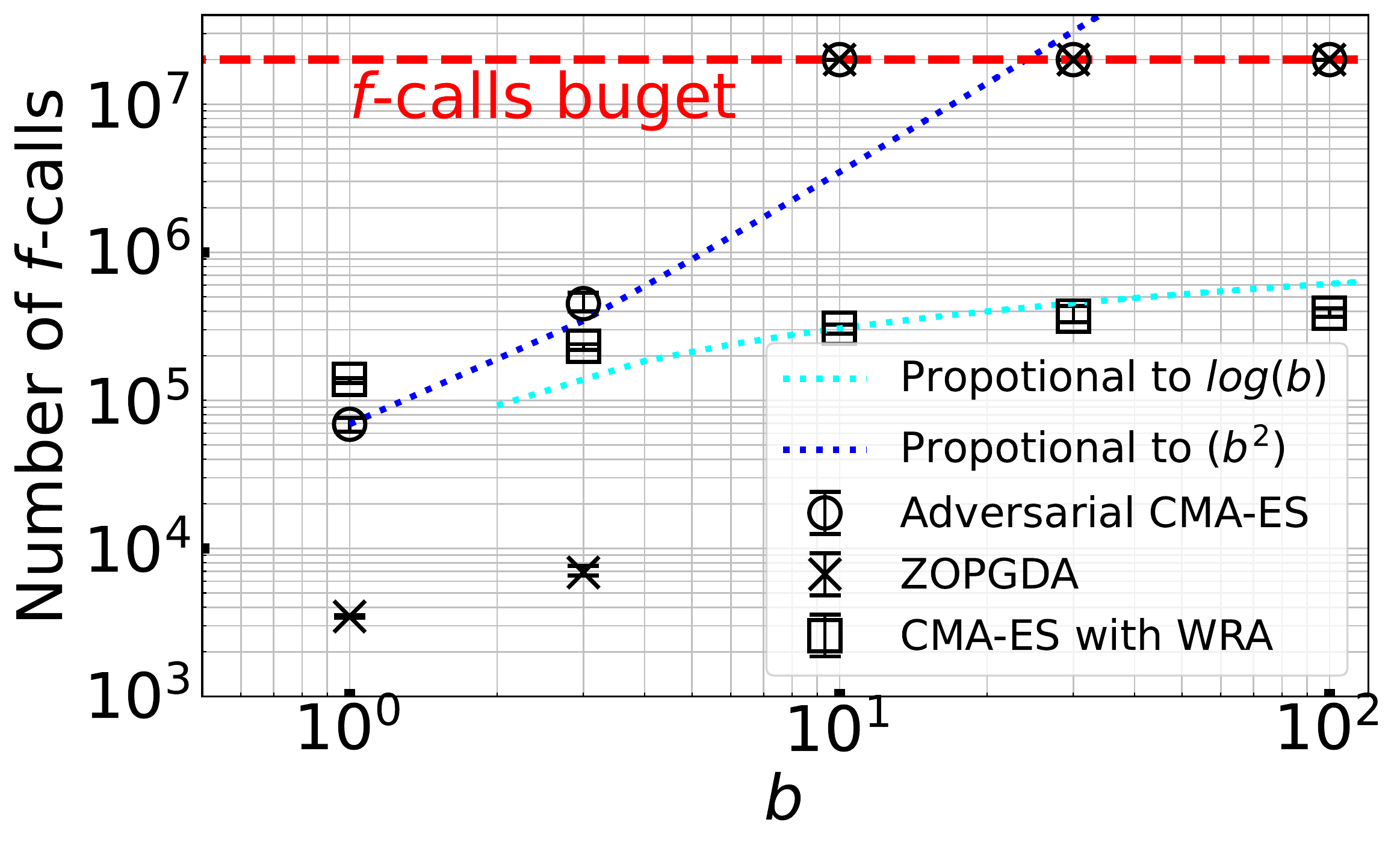}%
\caption{Comparison result among \amacma{}, \zopgda{}, and \acma{} at $b \in \{1, 3, 10, 30, 100 \}$ on $f_5$. Mean and standard deviation of the number of $f$-calls until successful convergence over 20 runs. \zopgda{} and \acma{} failed to converge at $b \in \{10, 30, 100 \}.$}
\label{fig:ex1}
\end{figure}

\providecommand{\exs}{0.25}
\begin{figure*}[t]
\centering%
\begin{subfigure}{\exs\textwidth}%
    \includegraphics[width=\textwidth]{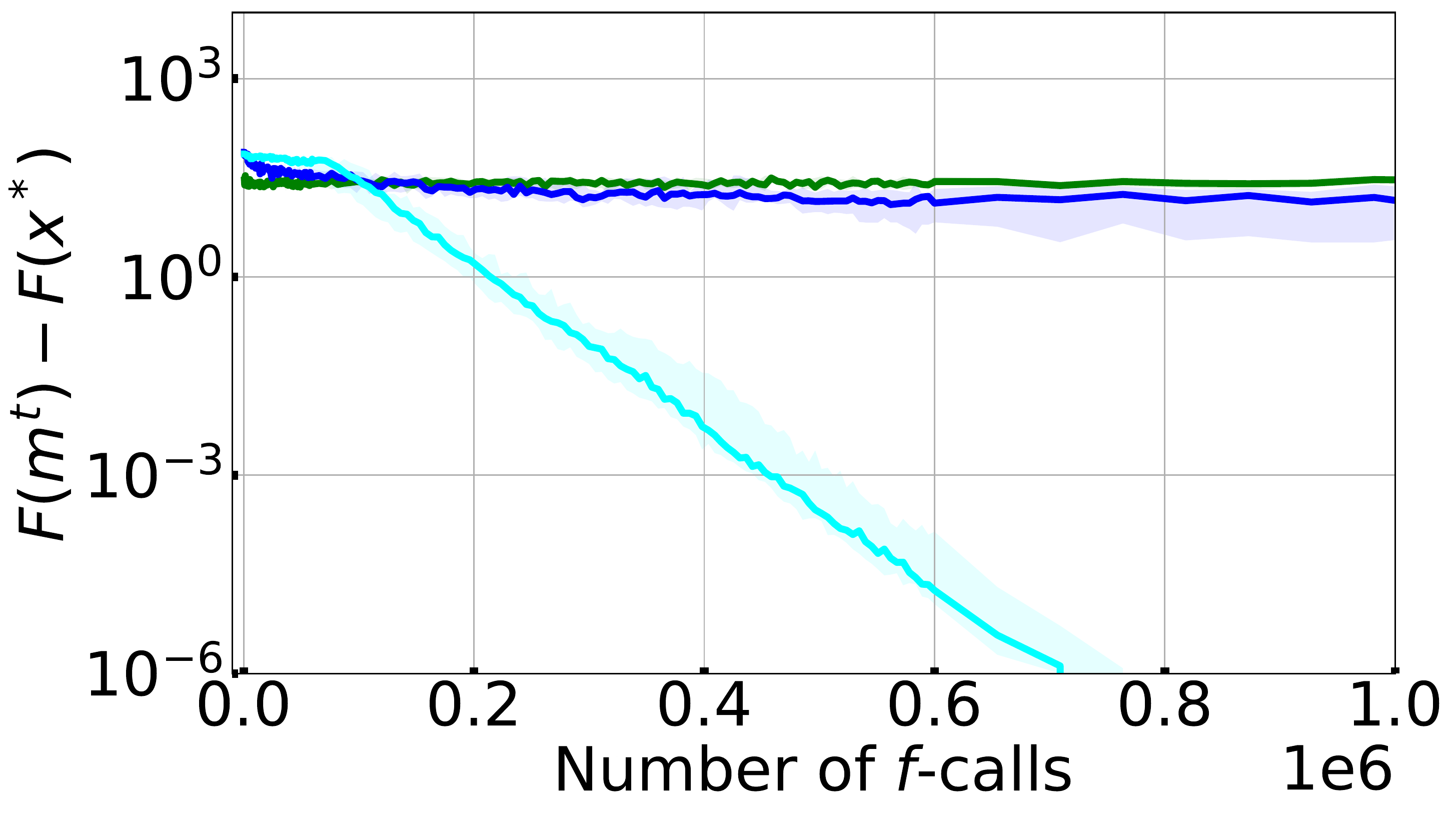}%
  \caption{$f_1$}%  
\end{subfigure}%
\begin{subfigure}{\exs\textwidth}%
    \includegraphics[width=\textwidth]{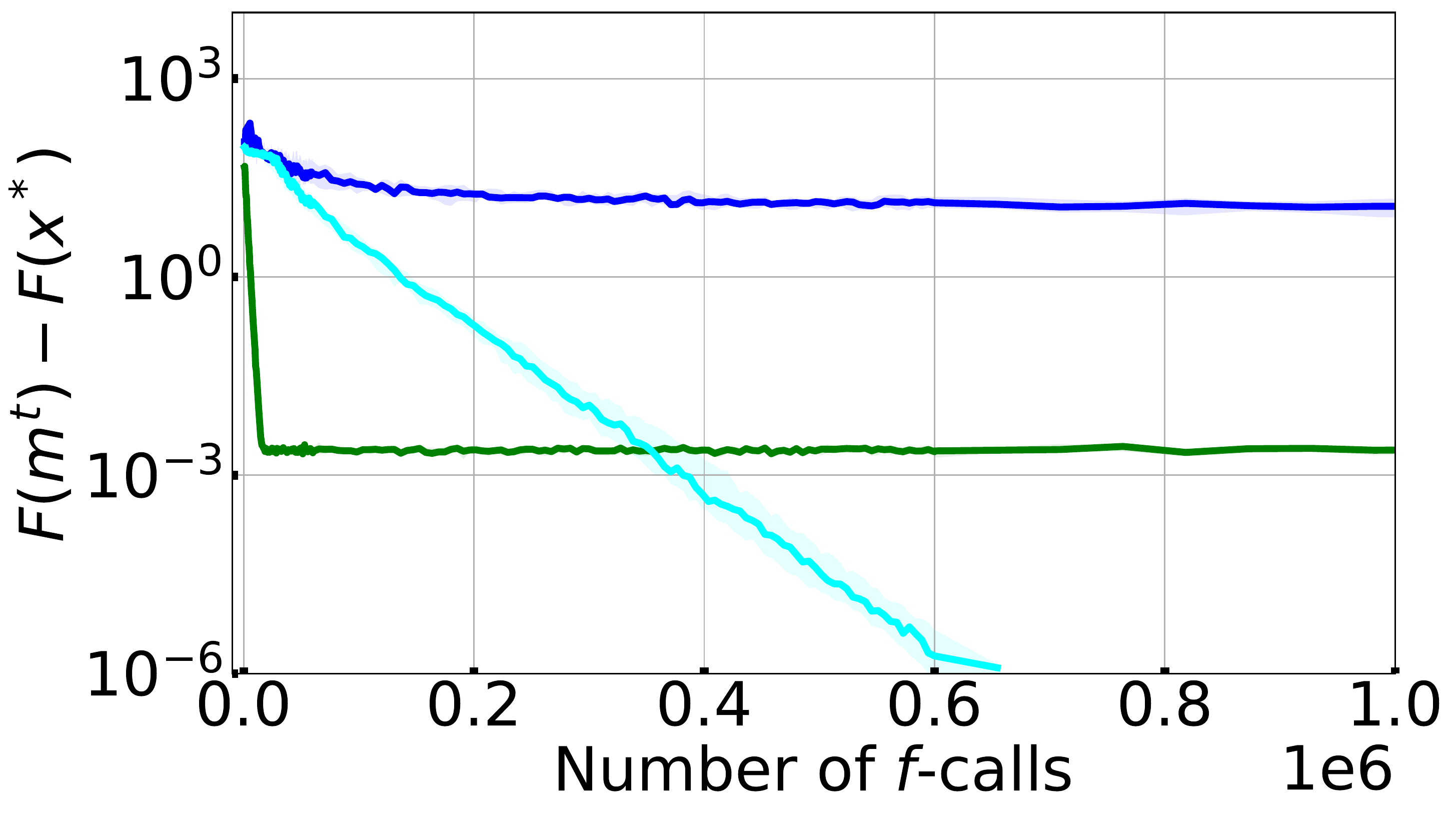}%
  \caption{$f_2$}%
\end{subfigure}%
\begin{subfigure}{\exs\textwidth}%
    \includegraphics[width=\textwidth]{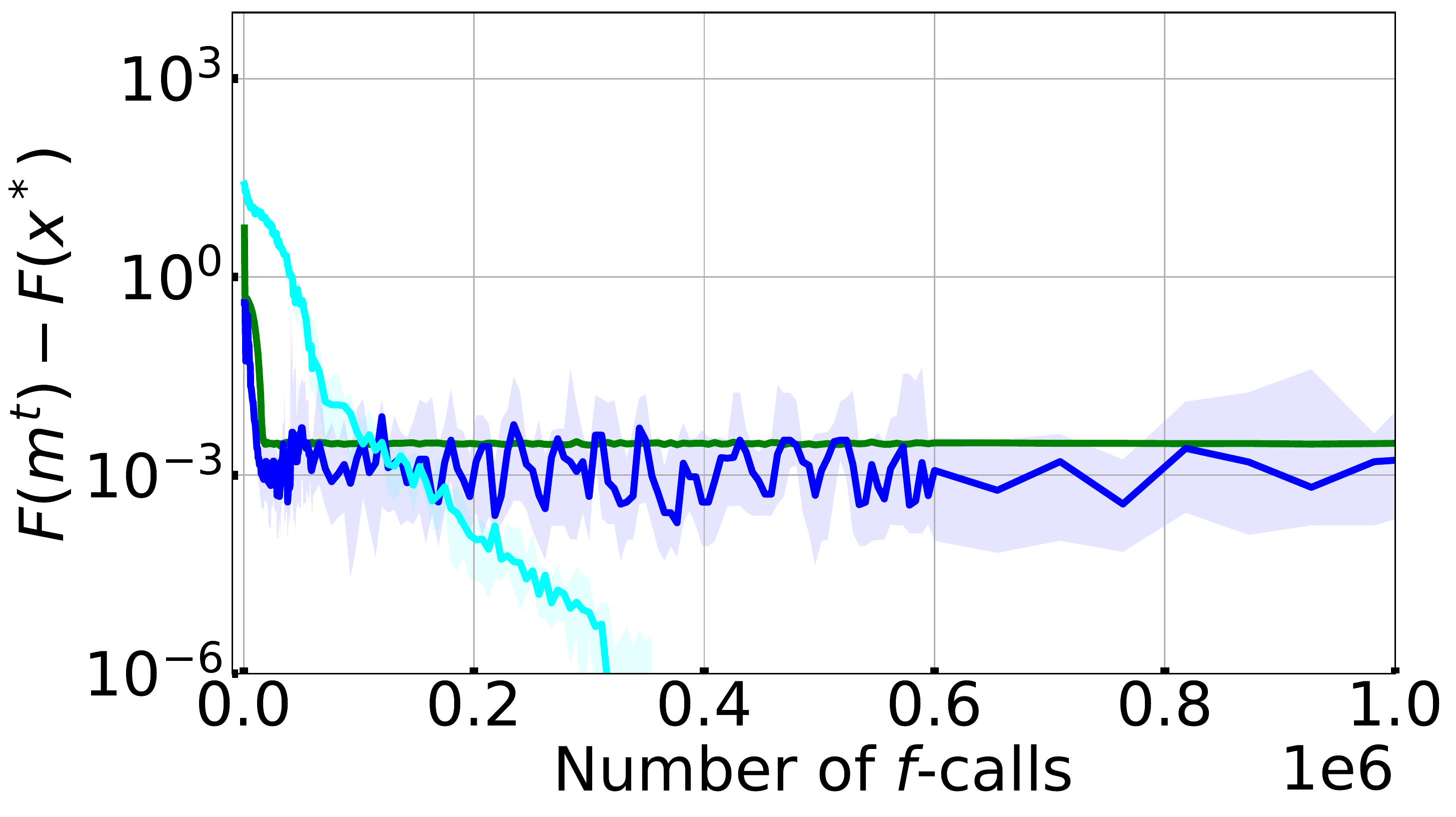}%
  \caption{$f_3$}%  
\end{subfigure}%
\begin{subfigure}{\exs\textwidth}%
    \includegraphics[width=\textwidth]{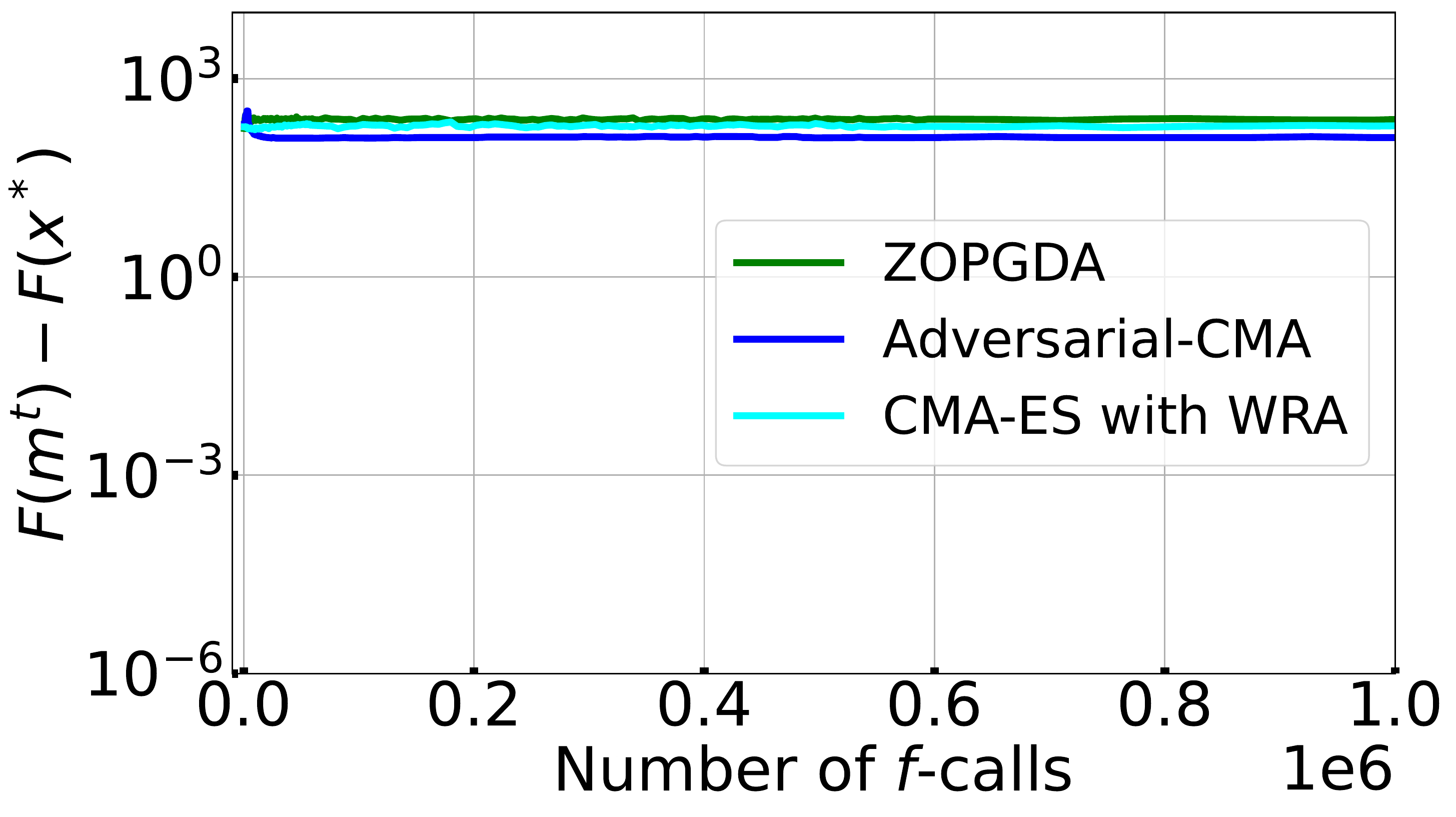}%
  \caption{$f_4$}%  
\end{subfigure}%
\\
\begin{subfigure}{\exs\textwidth}%
    \includegraphics[width=\textwidth]{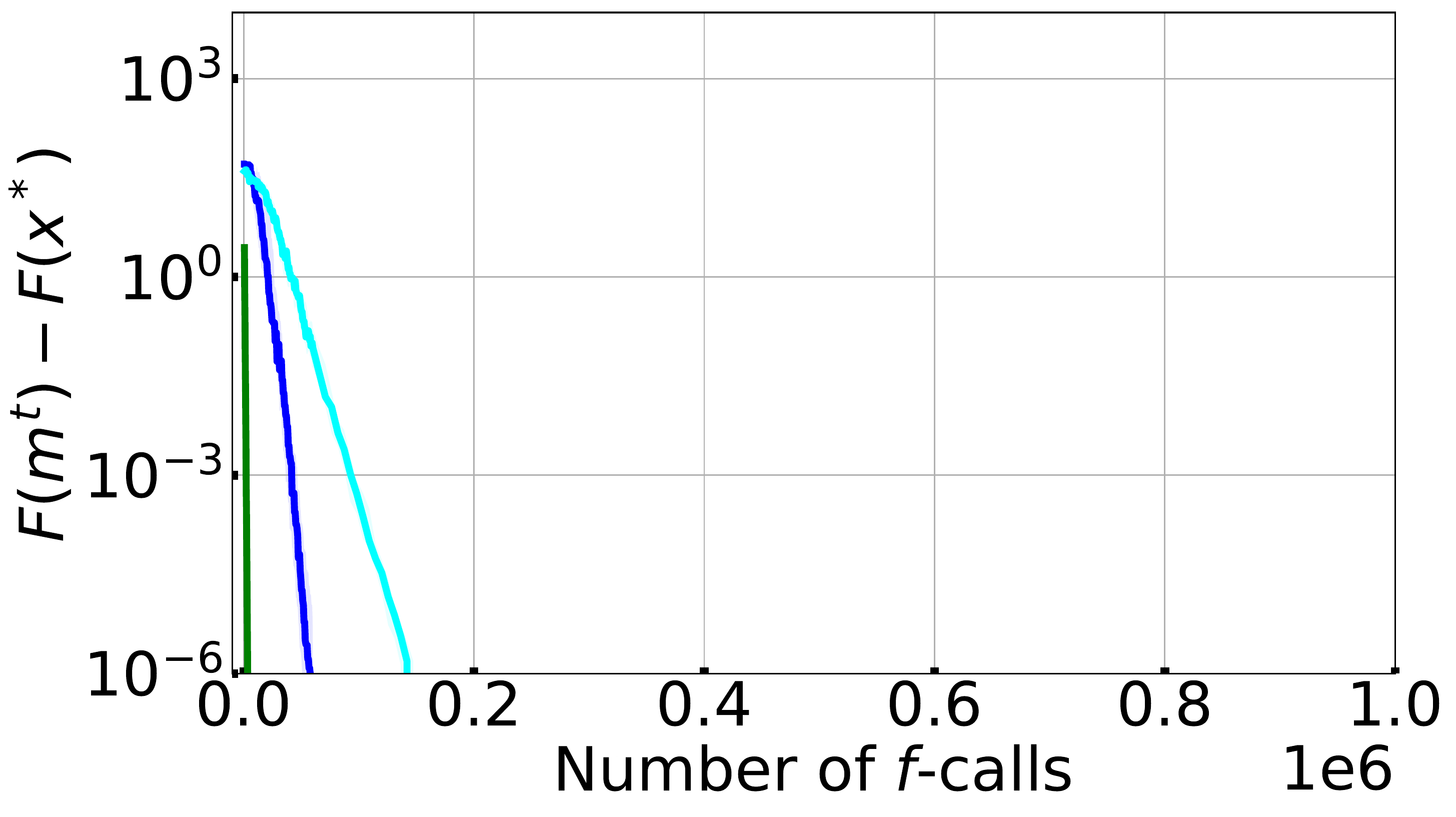}%
    \caption{$f_5$}%
\end{subfigure}%
\begin{subfigure}{\exs\textwidth}%
    \includegraphics[width=\textwidth]{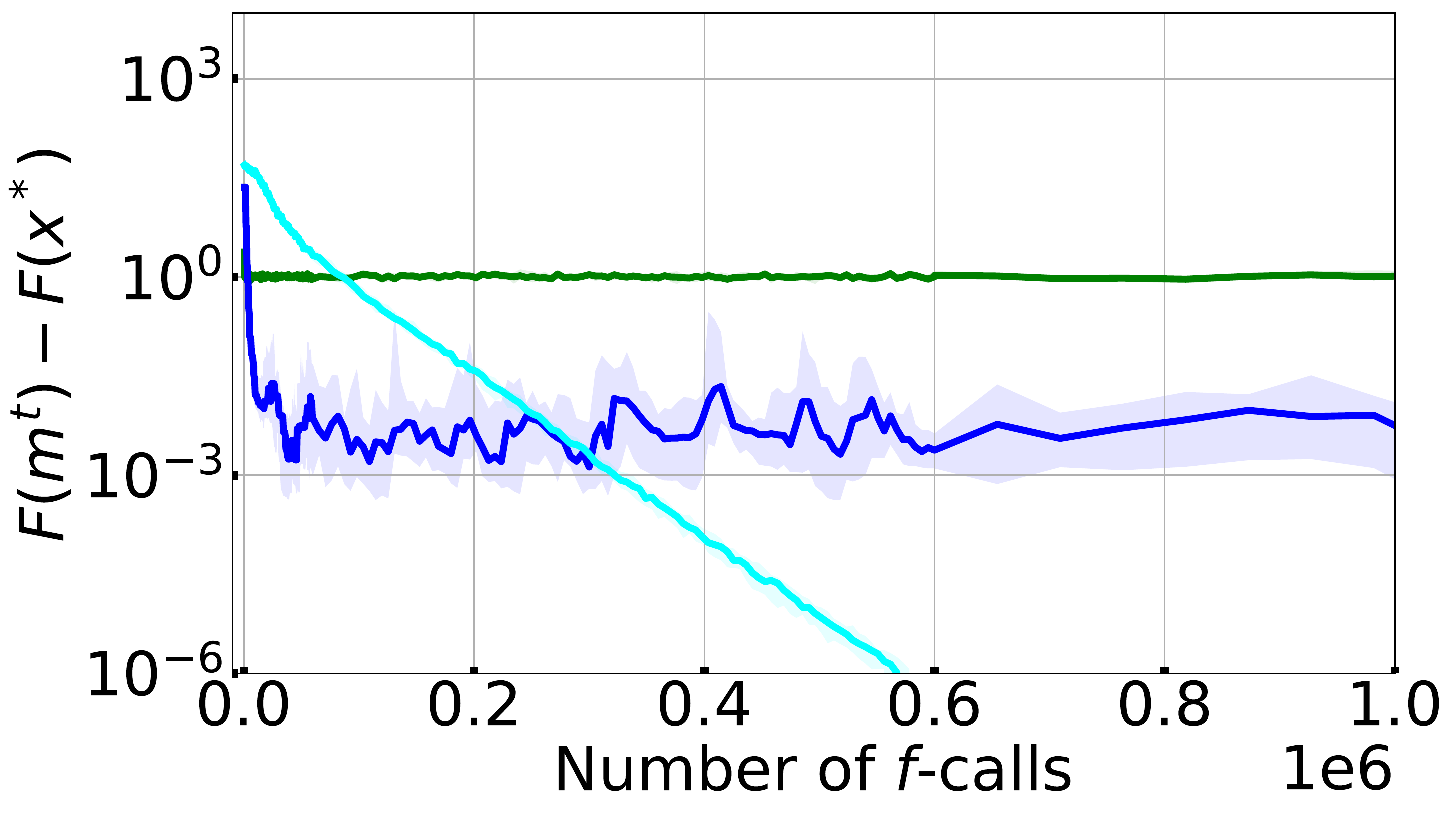}%
  \caption{$f_6$}%
\end{subfigure}%
\begin{subfigure}{\exs\textwidth}%
    \includegraphics[width=\textwidth]{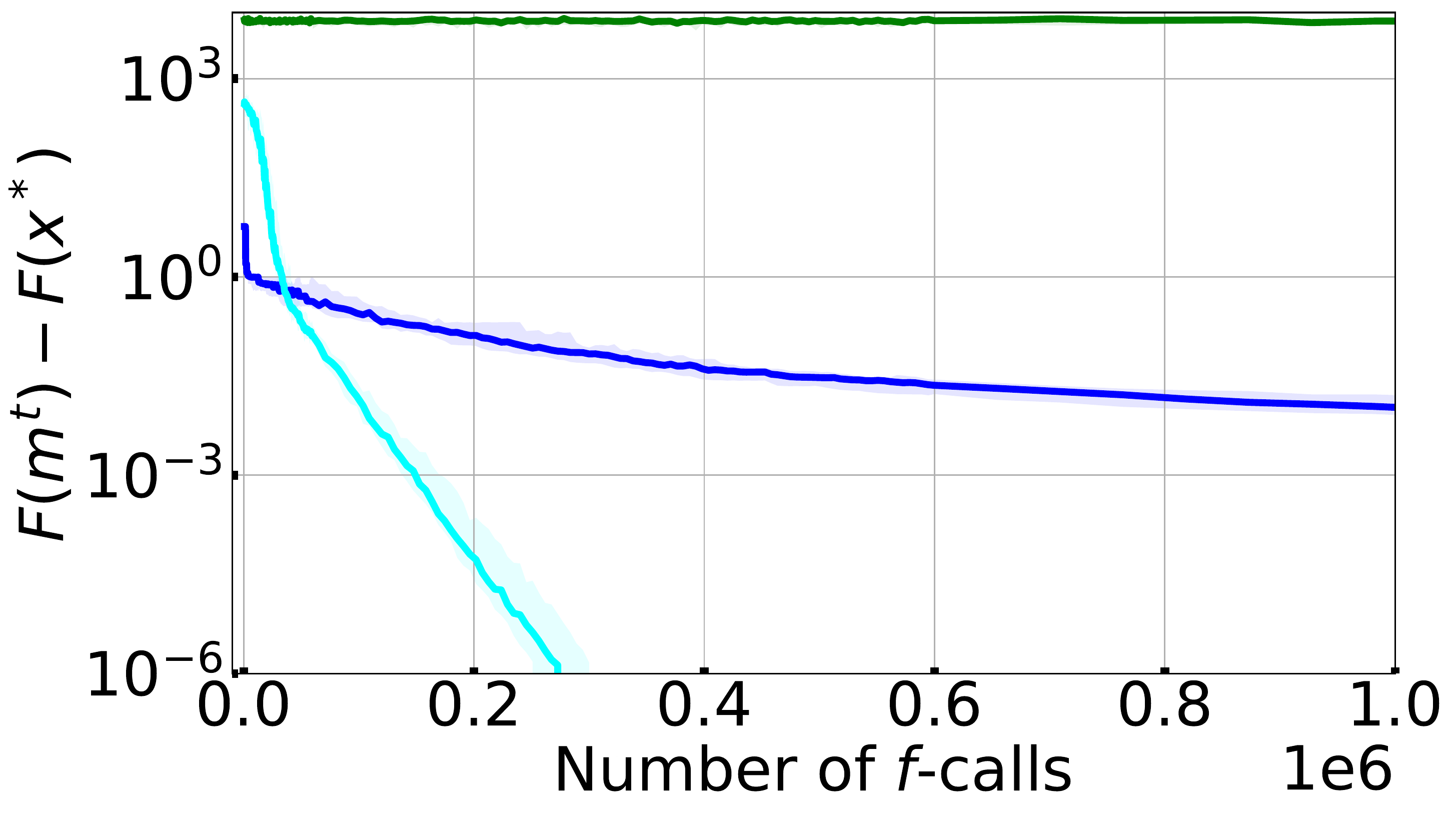}%
  \caption{$f_7$}%
\end{subfigure}%
\begin{subfigure}{\exs\textwidth}%
    \includegraphics[width=\textwidth]{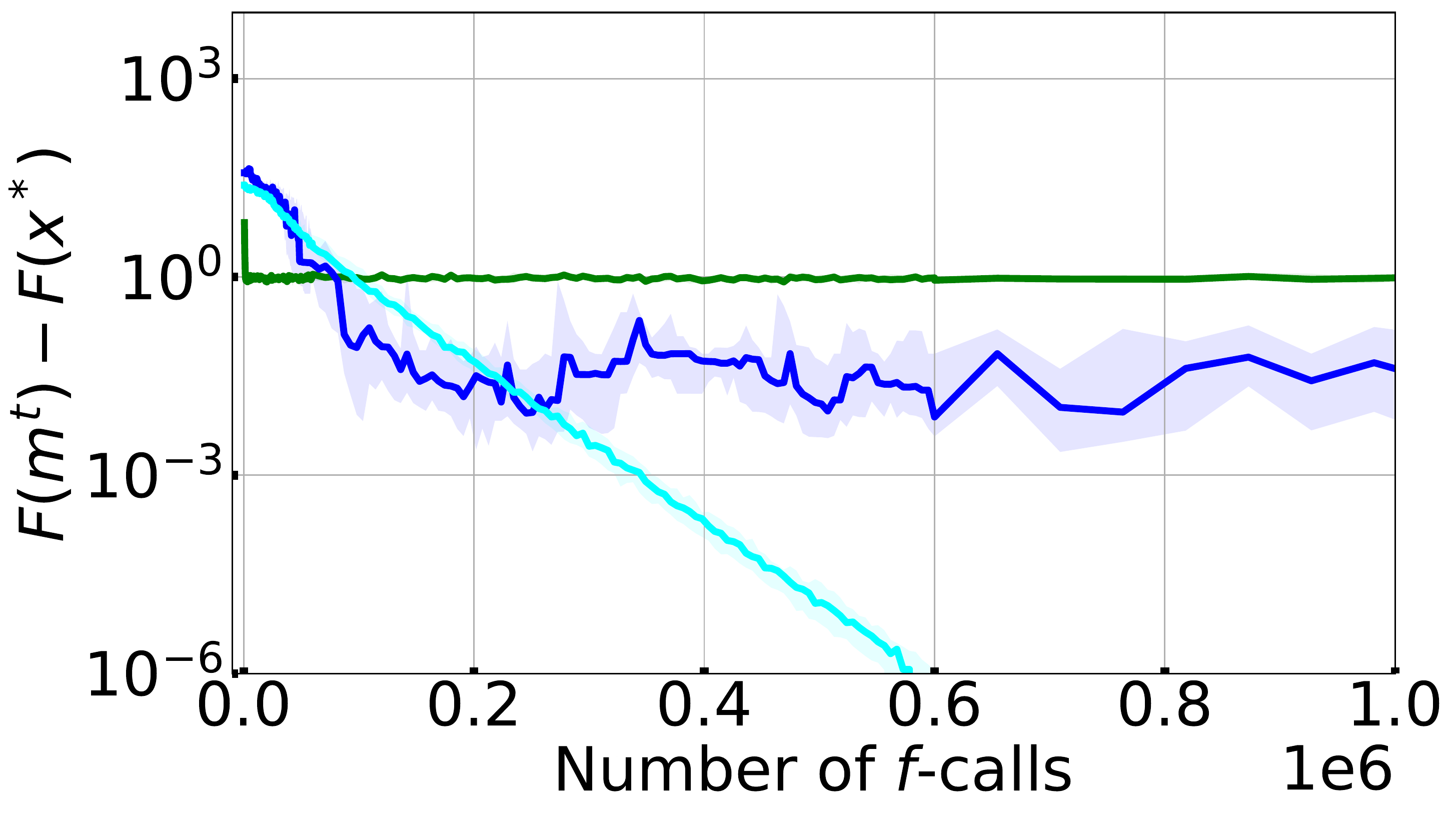}%
  \caption{$f_8$}%  
\end{subfigure}%
%\begin{subfigure}{0.15\textwidth}
%    \includegraphics[width=\textwidth]{./figures/legend.pdf}%
%\end{subfigure}
\caption{Gap $\abs{F(m^t) - F(x^*)}$ with the number of $f$-calls at $b=1$ on $f_1$--$f_8$.
The solid line represents the median (50 percentile) over 20 runs. The shaded area represents the interquartile range ($25$--$75$ percentile) over 20 runs.}%
\label{fig:ex2}
\end{figure*}
 \begin{figure*}[t]
\centering
\begin{subfigure}{\exss\textwidth}%
    \includegraphics[width=\textwidth]{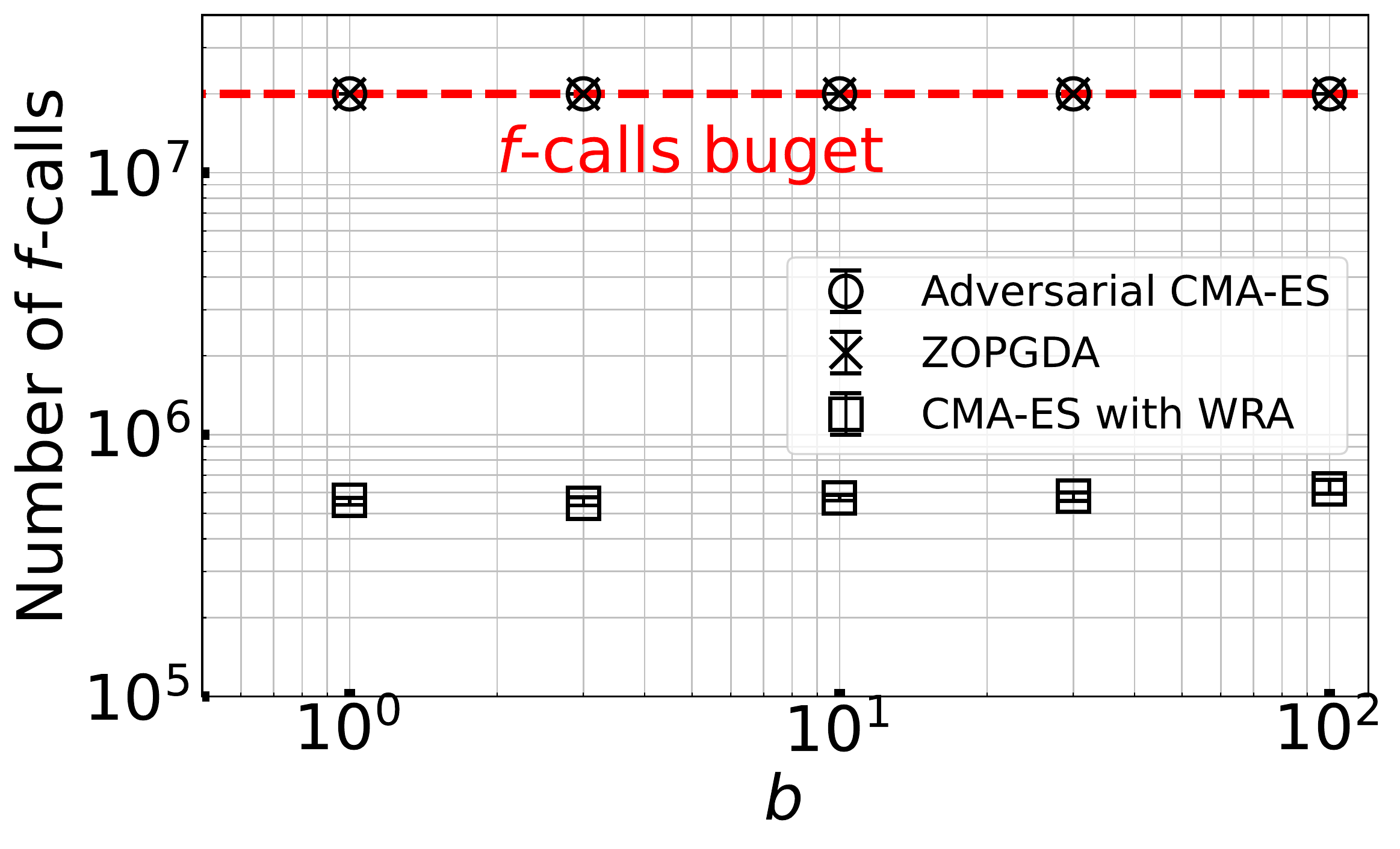}%
  \caption{$f_6$}%
\end{subfigure}%
\begin{subfigure}{\exss\textwidth}%
    \includegraphics[width=\textwidth]{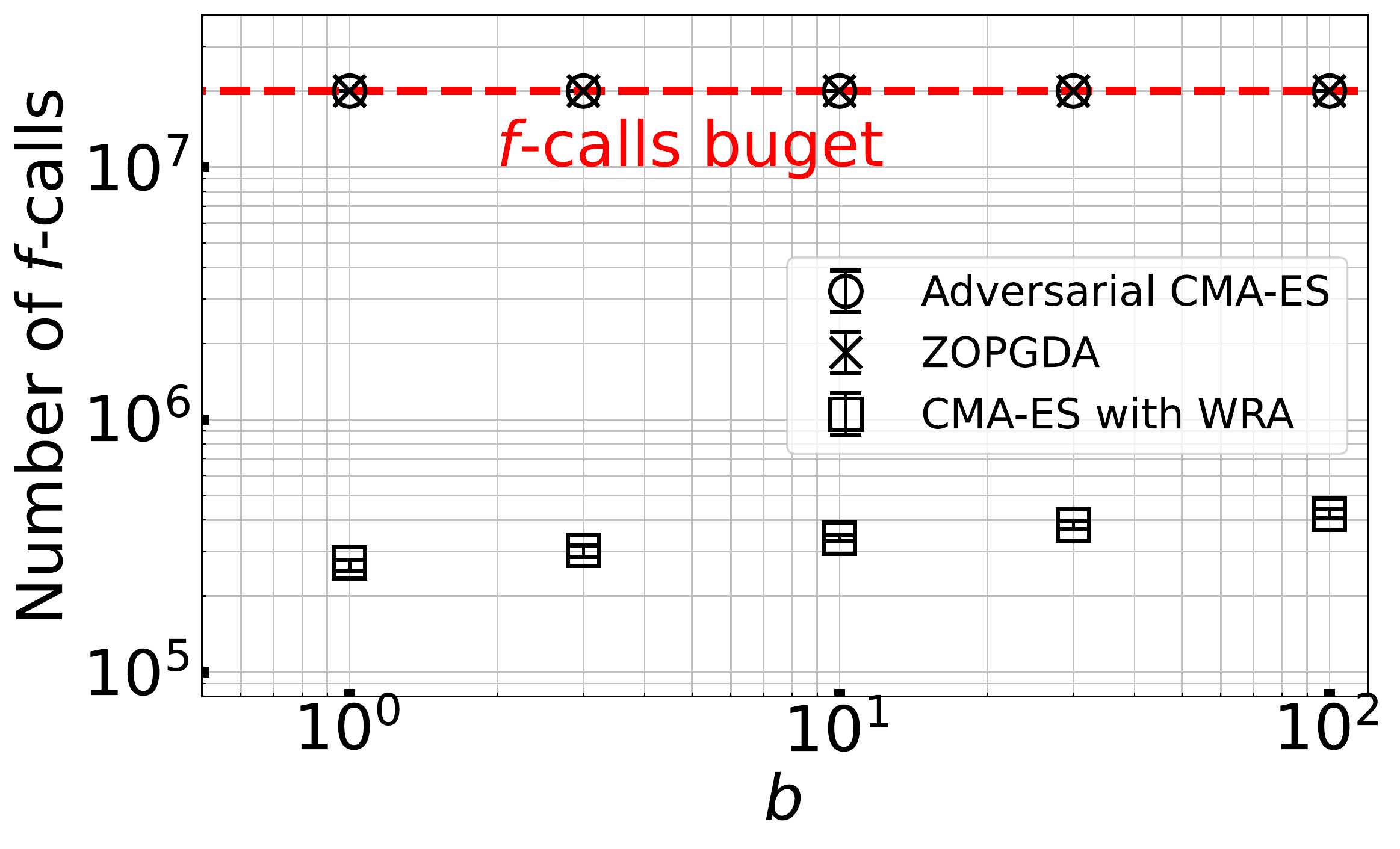}%
  \caption{$f_7$}%  
\end{subfigure}%
\begin{subfigure}{\exss\textwidth}%
    \includegraphics[width=\textwidth]{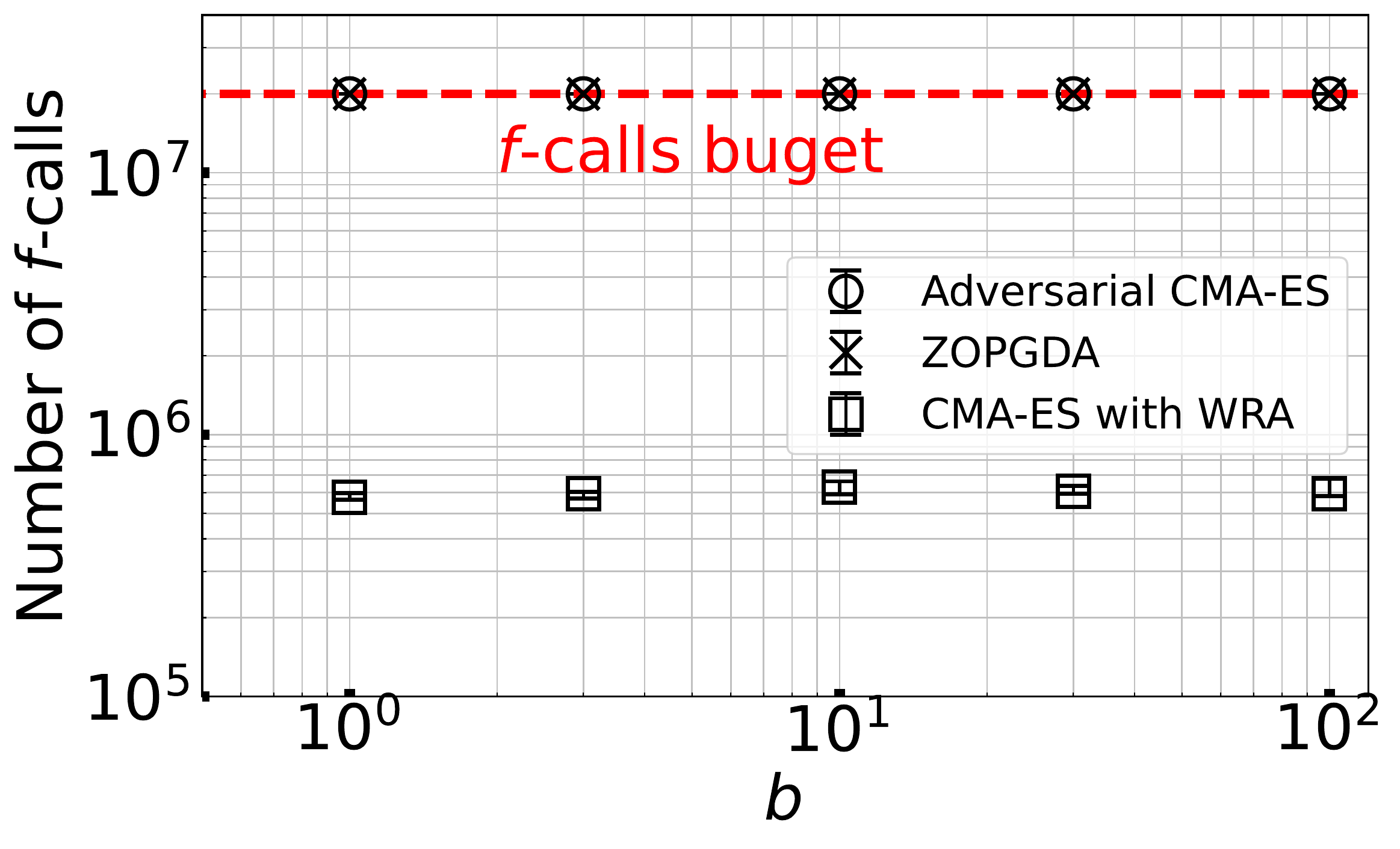}%
  \caption{$f_8$}%
\end{subfigure}%
\caption{Comparison result among \amacma{}, \zopgda{} and \acma{} at $b=\{1, 3, 10, 30, 100 \}$ on $f_6$--$f_8$. The mean and standard deviation of the number of $f$-calls until successful convergence over 20 runs.}
\label{fig:ex3}
\end{figure*}

\subsection{Experiment 2}\label{sec:ex2}

To verify hypothesis (b), we applied three approaches to $f_1$--$f_4$ and $f_6$--$f_8$. We set $b=1$ for $f_6$--$f_8$.   

The results of the experiment are shown in \Cref{fig:ex2}.
Except for the trials on $f_4$, \amacma{} achieved successful convergence in all trials.
Nevertheless, the existing approaches failed to determine the optimal solution in all trials. 
This was because the objective function in  the neighborhood of the optimal solution was not a Lipschitz smooth and strongly convex--concave function.

We discuss the results of \amacma{} on $f_4$. 
Let us consider the objective function $h(y)=f(x,y)$ for $y$ on a solution candidate $x \in \mathbb{X}$. This objective function $h(y)$ has the local optimal solution on the boundary of the search domain. Therefore, the objective function $h(y)$ has $2^n$ local optimal solutions, and it is considered a multi-modal function with a weak structure.
Such an objective function is difficult to optimize using any  currently proposed algorithm \cite{bbobresult}.
Therefore, \wra{} failed to approximate the worst-case ranking. Thus, CMA-ES could not converge to $x^*$ because it optimized for a function that differed significantly from $F$.

\subsection{Experiment 3}\label{sec:ex3}

To investigate the influence of the coefficient of the interaction term in the objective function, we applied three approaches to $f_6$--$f_8$ with $b \in \{1,3,10,30,100\}$. 

Experimental results are shown in \Cref{fig:ex3}.
We can confirm that \amacma{} could achieve successful convergence in all trials.
However, the existing approaches, \acma{} and \zopgda{}, failed to converge to the optimal solution in any trials on $f_6$--$f_8$. This is as verified in \Cref{sec:ex2}.

For \amacma{} on $f_6$--$f_8$, the number of $f$-calls required for convergence did not significantly change at various $b$. Even for the results on $f_7$, where the number of $f$-calls changed the most, the ratio of the number of $f$-calls at $b=1$ and $b=100$ was approximately two. The results of the existing approaches in \Cref{fig:ex1} suggest that the number of $f$-calls increased proportionally to $1+b^2 \approx 10^4$, implying that the factor of two can be considered as small.

\section{Discussion on the effect of the interaction term}\label{sec:discussion}

We discuss the effect of the interaction term of convex--concave problems on the number of required $f$-calls.
As observed in \Cref{fig:ex1} and \Cref{fig:ex3}, for $f_5, \dots, f_8$, the number of $f$-calls until \amacma{} reaches the target threshold of the worst-case objective function $F$ and scales with the coefficient $b$ of the interaction term $x^T y$ in the order of $O(\log(b))$.
Here, we provide an insightful but shallow analysis to describe this scaling.

We limit our attention to $f_5$ on an unbounded domain. 
The worst-case objective function is $F(x) = \frac{1 + b^2}{2}\norm{x}^2$. 

Moreover, we assume that CMA-ES converges linearly on an arbitrary convex quadratic objective function, that is, the number of $f$-calls that CMA-ES performs to reach $\{x : \norm{x - x^*} \leq \epsilon \cdot \norm{m^{(0)} - x^*}\}$ is $O(\log(\norm{m^{(0)} - x^*}/\epsilon))$, where $x^*$ is the optimal solution of the objective function. 
Although no rigorous runtime analysis has been conducted for CMA-ES, a variant of CMA-ES, namely, the (1+1)-ES, exhibited a linear convergence on strongly convex  functions with Lipschitz continuous gradients \cite{Morinaga2019foga}. 
Moreover, we empirically observe that CMA-ES geometrically approaches the optimal solution. 

Under this assumption, if CMA-ES is used to solve the worst-case objective function $F$ directly, the number of the worst-case objective function calls to reach $\{x : \norm{x - x^*} \leq \epsilon \cdot \norm{m^{(0)} - x^*}\}$, where $x^* = \argmin_x F(x)$, is $O(\log(\norm{m^{(0)} - x^*}/\epsilon))$.
The worst-case objective function is approximated to ensure that Kendall's rank correlation coefficient between the true values $\{F(x_i)\}$ and their approximated value $\{\hat{F}(x_i)\}$ is sufficiently high. Therefore, we expect that \amacma{} behaves similarly to  CMA-ES solving $F$ directly \cite{Akimoto2022surrogate}.
Therefore, we anticipate that \amacma{} must approximate $O(\log(\norm{m^{(0)} - x^*}/\epsilon))$ worst-case objective function values during the optimization. 

We now estimate the number of $f$-calls for each $F$ approximation. Because $F(x) \propto \norm{x}^2$, the covariance matrix of the upper-level CMA-ES is expected to be adapted as $\Sigma \approx \sigma^2 \cdot I$, where $\sigma^2 > 0$ is a scalar. The $\mathcal{L}_2$ distance between the worst-case objective function values of two candidate solutions $x_1$ and $x_2$ generated independently from $\mathcal{N}(m, \Sigma)$ is $\E[(F(x_1) - F(x_2))^2]^{1/2} = (1 + b^2) (2\Tr(\Sigma^2))^{1/2} \approx (2d)^{1/2} (1 + b^2)\sigma^2$. 
The approximated worst-case objective function $\tilde{F}$ is required to have a sufficiently high Kendall's rank correlation coefficient with the true value $F$ under the current search distribution $\mathcal{N}(m, \Sigma)$. Therefore,  it is assumed that the comparison $\tilde{F}(x_1) \lesseqgtr \tilde{F}(x_2)$ of two solutions generated from the current distribution provides a true comparison $F(x_1) \lesseqgtr F(x_2)$ with high probability.
Therefore, the worst-case objective function values need to be approximated with precision $\abs{F(x) - \tilde{F}(x)} \leq c^2 \cdot (1 + b^2) \cdot \sigma^2$ for some constant $c > 0$.
Observing that
\begin{align*}
    \MoveEqLeft[2]F(x) - \tilde{F}(x) 
    = f(x, \hat{y}(x)) - f(x, \tilde{y}) \\
    &= b x^\T(\hat{y}(x)- \tilde{y}) - (\norm{\hat{y}(x)}^2- \norm{\tilde{y}}^2) / 2\\
    &= \hat{y}(x)^\T(\hat{y}(x)- \tilde{y}) - (\norm{\hat{y}(x)}^2- \norm{\tilde{y}}^2) / 2
    = \norm{\hat{y}(x) - \tilde{y}}^2 / 2,
\end{align*}
the aforementioned condition reads $\norm{\hat{y}(x) - \tilde{y}} \leq c \cdot (1 + b^2)^{1/2} \sigma$.
The runtime to find such a $\tilde{y}$ using CMA-ES is $O\left(\log \left(\frac{\norm{y^{(0)} - \hat{y}(x)}}{c \cdot (1 + b^2)^{1/2} \sigma}\right)\right)$.
Notably, $\hat{y}(x) = b x$ and $\tilde{y}^{(0)}$ is a near-optimal solution to $\max_y f(x, y)$ for a solution $x$ generated in a previous iteration. As the distribution parameters of the upper-level CMA-ES do not change rapidly, $m$ and $\Sigma$ remain from the last iteration. 
Thus, both $\hat{y}(x)$ and $\tilde{y}^{(0)}$ can be considered $\mathcal{N}(b m, b^2 \Sigma)$-distributed. 
Their expected squared distance is then $b^2 \Tr(\Sigma) \approx d b^2\sigma^2$. 
Therefore, we estimate $\norm{y^{(0)} - \hat{y}(x)} \in O(b \sigma)$.
Hence, the runtime to find such a $\tilde{y}$ is $O\left(\log \left(\frac{b}{c\cdot (1 + b^2)^{1/2}}\right)\right)$. 
That is, the number of $f$-calls required to approximate the worst-case objective function value for each $x$ remains constant order over time.

Generally, we obtain the estimated number of $f$-calls until \amacma{} reaches the $\epsilon$-optimal solution to the worst-case objective function as follows:
\begin{equation}
    O\left( \log \left(\frac{b}{c \cdot (1 + b^2)^{1/2}}\right) \cdot \log\left( \frac{\norm{m^{(0)} - x^*}}{\epsilon}\right)\right) .
\end{equation}
For $b \leq 1$, the first term scales as $\log(b)$. 
However, as $b \to \infty$, the first term approaches a constant. 

\section{Conclusion}

This study focused on min--max continuous optimization problems whose objective function is a black-box.
We addressed the following challenges of the existing approaches, \acma{} and \zopgda{}. 
(I) The number of $f$-calls required to reach convergence depends largely on the interaction term $x^T y$ of the objective function.
(II) The objective function in the neighborhood of the optimal solution needs to be a Lipschitz smooth and strongly convex--concave function for convergence.

Our contributions are as follows. 
(A) We proposed a new approach (CMA-ES with a worst-case ranking approximation: \amacma{}) to address Difficulty (I) and (II). \amacma{} works because \wra{} estimates the ranking of the solution candidates on the worst-case function, and CMA-ES searches for the optimal solution using the estimated ranking information. 
(B) Numerical experiments on the strongly convex--concave function showed that \amacma{} improved the scalability of the number of $f$-calls against the coefficient $b$ multiplied by $x^T y$. The number of $f$-calls resulting from \amacma{} scaled to approximately $\log(b)$, whereas that of the existing approaches was $b^2$. 
(C) To ensure that \amacma{} addresses Difficulty~(II),  we applied \amacma{} to test problems whose objective function was not limited to being Lipschitz smooth and strongly convex--concave in the neighborhood of the optimal solution. The experimental results showed that only \amacma{}, among the compared approaches, could converge to the optimal solution. However, \amacma{} could not converge to the optimal solution when the worst-case ranking could not be estimated properly, for example, when the objective function was a multi-modal function with a weak structure. 
(D) Additionally, we confirmed that the number of $f$-calls performed by \amacma{} was not significantly affected by changing the coefficient $b$ multiplied by $x^T y$ on the objective functions that are not limited to being a Lipschitz smooth and strongly convex--concave function in the neighborhood of the optimal solution.

% The reduction of the number of $f$-calls of \amacma{} will be treated as a future work. For example, $f_1$ and $f_2$ required approximately  $15,0000,000$ $f$-calls, which is unrealistic for real-world application. To reduce this, one approach is to change the optimization algorithm that approximately solves the internal maximization problem from CMA-ES to some subgradient approaches.
% In addition, sensitivity analysis on $c_{\max}$ and $\tau_\mathrm{threshold}$ will also be  treated in future works. 
The limitations of this study are the lack of a theoretical analysis of the proposed approach and empirical evaluation on the scaling of the number of $f$-calls about the dimension $m$ and $n$ on broader class of functions. 
Moreover, the successful convergence of the proposed approach was not clearly identified. The sensitivity analysis to the hyper-parameters of \wra{}, $c_{\max}$ and $\tau_\mathrm{threshold}$, is yet to be performed.
Future work on the proposed approach should include more theoretical and empirical analyses. 
Compared with the existing approaches, \zopgda{} and \acma{}, \amacma{} requires significantly more $f$-calls if the objective function is strongly convex-concave and Lipschitz continuous and the effect of the interaction term, $H_{x,y}$, is relatively small. This is, therefore, a limitation of the proposed approach.

\begin{acks}
This paper is partially supported by JSPS KAKENHI Grant Number 19H04179.
\end{acks}

% \bibliographystyle{ACM-Reference-Format}
% \bibliography{./thebibliography.bib}


\end{document}